\documentclass[letterpaper, 10 pt, conference]{ieeeconf}  

\IEEEoverridecommandlockouts 

\usepackage{epsfig}
\usepackage{graphicx}
\usepackage{amsmath}
\usepackage{amssymb}
\usepackage{tabularx}
\usepackage{multirow}
\usepackage{caption}
\usepackage[export]{adjustbox}
\usepackage{subfig}
\usepackage{tabu}
\usepackage{comment}
\usepackage{ctable}
\usepackage{boldline}
\usepackage{pifont}
\usepackage{float}
\usepackage{tablefootnote}
\usepackage{footnote}
\usepackage{bm}

\usepackage[pagebackref=true,breaklinks=true,colorlinks,bookmarks=false]{hyperref}

\makeatletter
\let\NAT@parse\undefined
\makeatother
\usepackage[numbers,sort&compress]{natbib}

\definecolor{viz_red}{RGB}{196, 0, 0}
\definecolor{viz_yellow}{RGB}{241, 194, 50}
\definecolor{viz_green}{RGB}{52, 183, 47}
\definecolor{viz_blue}{RGB}{17, 85, 204}

\title{\LARGE \bf
KDFNet: Learning Keypoint Distance Field for 6D Object \\ Pose Estimation
}

\author{Xingyu Liu, Shun Iwase, Kris M. Kitani
\thanks{The authors are with the Robotics Institute, Carnegie Mellon University, 5000 Forbes Ave, Pittsburgh, PA 15213, United States. Contact: {\tt\small \{xingyul3, siwase, kkitani\}@cs.cmu.edu }}
}

\begin{document}

\maketitle
\thispagestyle{empty}
\pagestyle{empty}

\begin{abstract}

We present KDFNet, a novel method for 6D object pose estimation from RGB images. 
To handle occlusion, many recent works have proposed to localize 2D keypoints through pixel-wise voting and solve a Perspective-n-Point (PnP) problem for pose estimation, which achieves leading performance.
However, such voting process is direction-based and cannot handle long and thin objects where the direction intersections cannot be robustly found.  
To address this problem, we propose a novel continuous representation called Keypoint Distance Field (KDF) for projected 2D keypoint locations.
Formulated as a 2D array, each element of the KDF stores the 2D Euclidean distance between the corresponding image pixel and a specified projected 2D keypoint.
We use a fully convolutional neural network to regress the KDF for each keypoint.
Using this KDF encoding of projected object keypoint locations, we propose to use a distance-based voting scheme to localize the keypoints by calculating circle intersections in a RANSAC fashion.
We validate the design choices of our framework by extensive ablation experiments.
Our proposed method achieves state-of-the-art performance on Occlusion LINEMOD dataset with an average ADD(-S) accuracy of 50.3\% and TOD dataset mug subset with an average ADD accuracy of 75.72\%.
Extensive experiments and visualizations demonstrate that the proposed method is able to robustly estimate the 6D pose in challenging scenarios including occlusion.

\end{abstract}

\vspace{1ex}

\section{Introduction}

Estimating the 6D pose of a rigid object, i.e. rotation and
translation in 3D space, is a core problem in computer vision and is crucial for applications such as robotic manipulation and augmented reality (AR).
We focus on the setting of 6D object pose estimation of a rigid object from RGB images where the 3D textured mesh model of the object is known.
Early attempts to this problem include direct pose regression using neural networks in an end-to-end fashion \cite{posenet, posecnn}.
Recently, 6D pose estimation methods that use object keypoints as intermediate representation have been successful and achieve leading performance on various benchmarks \cite{6dof:kp,keypose,pvnet,hybridpose}.
By definition, keypoints are 3D points attached to an object model and are usually a subset of the object surface points.
In keypoint-based methods, 2D projections of 3D object keypoints or centers are first located on the image and then the 6D pose can be recovered from such 2D-3D correspondences by solving a Perspective-n-Point (PnP) problem.

There are mainly two types of methods for localizing 2D keypoints: heatmap-based \cite{6dof:kp,keypose} and voting-based \cite{posecnn,pvnet,hybridpose}.
Heatmap-based methods predict probability heatmaps of a keypoint over the image and localize it through an integral operation with the image coordinate map.
Though heatmap-based method achieves strong performance on problems such as human pose estimation \cite{integralnet}, it is known to be vulnerable to occlusion, because the features of the occluder near the keypoint location can significantly affect the predicted probability map.
In voting-based methods, the visible parts of the object hallucinate and vote for the 2D locations of the invisible keypoints \cite{pvnet}.
The training objectives of the votings are independent of the occluders.
Therefore, compared to heatmap-based methods, voting-based methods are more robust to occlusions and achieve stronger performance on object pose estimation benchmarks where occlusions are important. 

The voting schemes of existing methods are direction-based, i.e. every object pixel predicts the 2D direction to the keypoints and the keypoint hypotheses are the intersections of the direction votes \cite{posecnn,pvnet,hybridpose}.
Direction-based voting methods are built upon an important assumption: the angles between the voting directions are large enough so that the keypoint hypotheses can be reliably found by computing the intersections of voted directions.
However, this assumption does not hold for long and thin objects where most voting directions concentrate in a small range and the keypoint hypotheses cannot be found or are extremely sensitive to noise in direction.

\begin{figure}[t]
\centering
\small
\subfloat[]{%
    \includegraphics[height=0.3\linewidth]{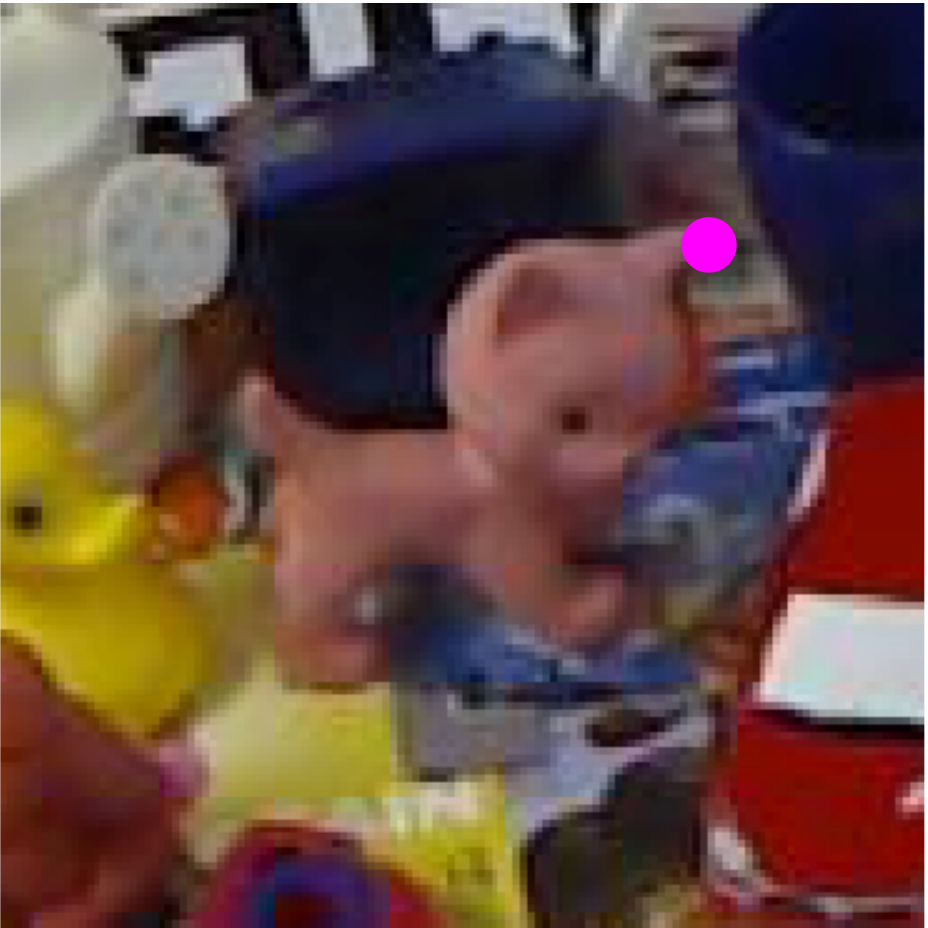}
    \label{fig:teaser:im}
}
\subfloat[]{%
    \includegraphics[height=0.3\linewidth]{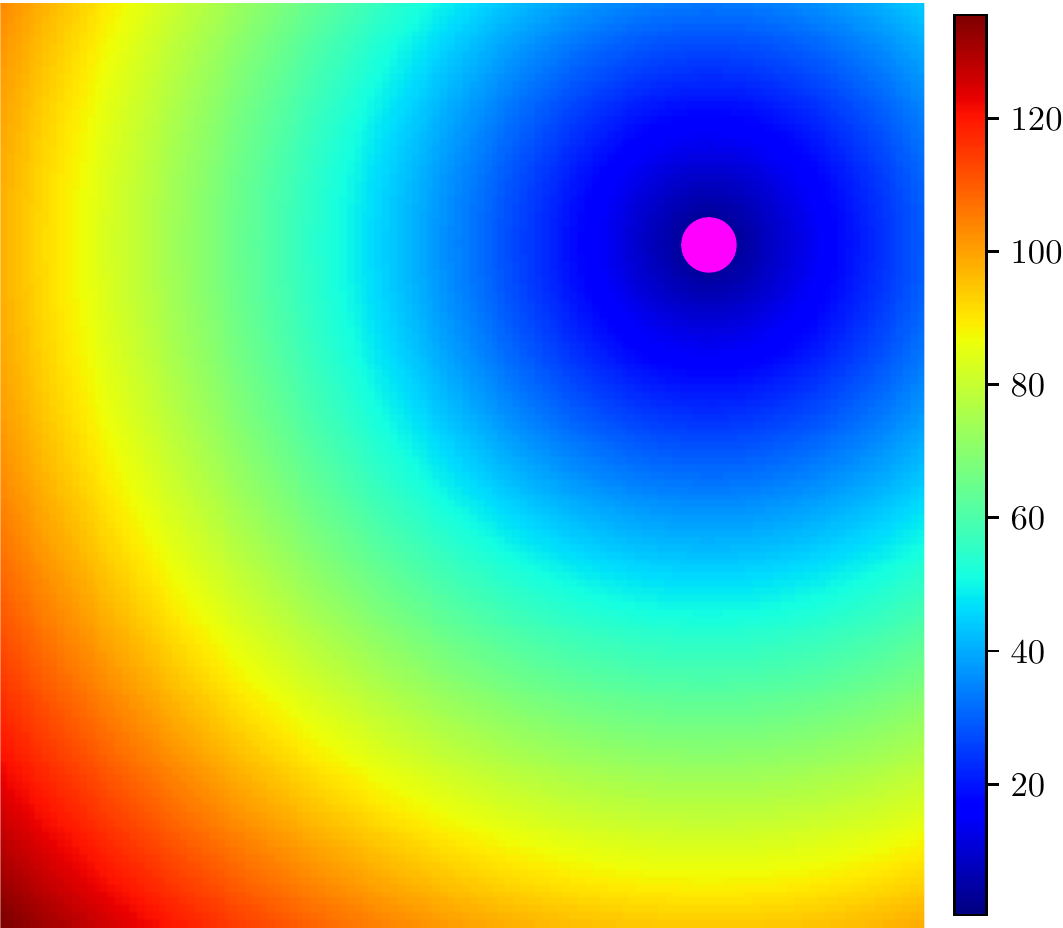}
    \label{fig:teaser:gt}
}
\subfloat[]{%
    \includegraphics[height=0.3\linewidth]{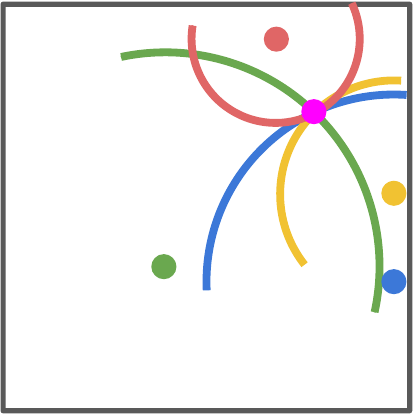}
    \label{fig:teaser:voting}
}
\caption{ (a) Our method estimates object 6D pose from RGB images by localizing object keypoints and solving a Perspective-n-Point (PnP) problem. (b) For every keypoint, our deep model regresses \textbf{Keypoint Distance Field (KDF)}, a novel 2D map representation for localizing projected keypoints. (c) To recover the 2D locations of the projected keypoints from regressed KDFs, we propose a \textbf{distance-based voting scheme} that includes calculating circle intersections in RANSAC fashion.
}
\label{fig:teaser}
\end{figure}

To address this problem, we propose a novel representation for object keypoint named \emph{Keypoint Distance Field (KDF)}.
Defined as a 2D map of the same spatial size as the RGB image, the KDF is the function of 2D Euclidean distances to a certain projected keypoint.
Given a perfect KDF, the 2D location of the projected keypoint can be easily recovered.
Note that KDF is able to represent keypoints that are invisible or even outside the image field of view.
Given $K$ keypoints, there are $K$ KDFs defined.

With the KDF representation, in this paper, we introduce a novel keypoint-based 6D object pose estimation framework named KDFNet.
The core of KDFNet is a fully convolutional neural network (CNN) that predicts the KDFs of the object keypoints through per-pixel regression.
To efficiently recover the 2D locations of the projected keypoints from the predicted KDFs, we propose a distance-based voting scheme.
The voting hypotheses are generated through circle intersections where the centers of the circles are the pixel voters and the radii of the circles are the predicted keypoint distance values.
The projected keypoints are the hypothesis with the maximum consensus of distance prediction among pixel voters.
The core idea is illustrated in Figure \ref{fig:teaser}.

We evaluated our framework on one of the most popular 6D pose estimation benchmarks, Occlusion LINEMOD \cite{occlusion:linemod}, and compare it against related baseline approaches. 
On Occlusion LINEMOD, our method achieves an accuracy of 50.3\%, significantly outperforms related baselines such as \cite{pvnet} and the current state-of-the-art HybridPose \cite{hybridpose}.
In addition, we evaluate the keypoint estimation of KDFNet on TOD \cite{keypose}, a stereo-RGB dataset for object pose estimation, and compare against previous stereo-RGB keypoint methods including KeyPose \cite{keypose}.
On TOD dataset, our method also achieves state-of-the-art results.

\begin{figure*}[t]
\centering
\small
\includegraphics[width=\linewidth]{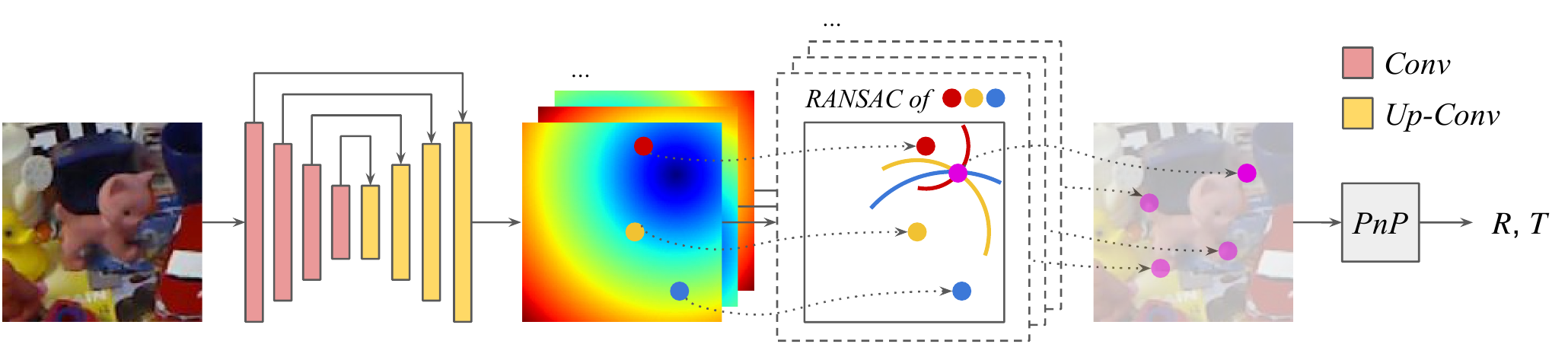}
\caption{\textbf{Overall Architecture of KDFNet}. 
We used a fully convolutional neural network to predict the KDFs defined by all keypoints of the objects. 
For every predicted KDF, the distance-based voting scheme randomly samples three positions \textcolor{viz_red}{ $\mathbf{p}_{i_1}$}, \textcolor{viz_yellow}{$\mathbf{p}_{i_2}$}, \textcolor{viz_blue}{$\mathbf{p}_{i_3}$} for $N$ times to generate keypoint hypotheses. 
The keypoints location are determined by selecting the hypothesis with the highest voting score defined by Equation \eqref{eq:vote:score}.
Finally, the 6D pose is calculated by solving a PnP problem from the predicted object keypoint locations. }
\label{fig:overall:arch}
\vspace{-2ex}
\end{figure*}

\section{Related Work}

\textbf{Learning Object Keypoints.}
Many previous works have explored deep learning methods for localizing 3D keypoints of an object \cite{keypointnet,6dof:kp,keypose,pvn3d,kundu2018object} or a human \cite{integralnet,monocular:3d:human:pose} from an RGB image to estimate their poses.
The core of these methods is to predict the probability heatmap for the 2D keypoint and localizing it through integral with coordinate map.
This idea is also used in 2D object detection where the corners of 2D object bounding boxes are formulated as keypoints and are localized through probability heatmaps \cite{centernet,cornernet}.
Besides heatmap, 2D direction field representation has also been proposed for localizing keypoints \cite{pvnet,posecnn}. 
Our method also uses keypoint 2D location as a bridge to 6D object pose.
We propose a novel distance field representation of keypoints and the associated voting scheme.

\textbf{Object Pose Estimation}
The earliest attempts for object pose estimation used CNN to directly regress 6D pose \cite{posenet,posecnn,ssd:6d}.
Rough 2D bounding boxes of the objects may be predicted first to localize the objects more accurately \cite{posecnn}.
However, directly estimating the 6D poses using CNN assumes the neural network can implicitly remember the images of the object in all possible 6D poses which is difficult and prone to occlusion and background clutter.
Instead, methods such as \cite{dpod,pix2pose} leverages object coordinate maps as the dense 2D-3D correspondence representation for 6D pose.
Compared to dense methods, keypoints are a more flexible representation and are used in \cite{pvnet,hybridpose,keypose,kundu2018object}. 
Our pose estimation method is keypoint-based and predicts distance maps to localize the keypoints.

\textbf{Voting for Understanding Objectness.} 
Voting has also been used in 3D object detection where objectness can be inferred from voting consensus \cite{votenet,implicit:shape:models,local:rgbd:patches,pose:hough:voting}. 
Specifically, direction-based voting methods have been adopted by previous works to robustly localize object centers \cite{posecnn} or object keypoints \cite{pvnet,pvn3d} on the images where the voting scores are based on the number of direction inliers.
Instead of calculating the mean of predictions, voting-based methods find the maximum consensus among predictions.
Therefore, voting-based methods are known to be robust to noise and occlusions. 
Our method also leverages voting for robustly detecting keypoints. 
Different from previous methods, our voting scheme is distance-based.

\section{Method}

Given an RGB image of a rigid object, the goal of our framework is to predict the rotation  $\mathbf{R}$ and translation $\mathbf{T}$ of the object.
One of the popular approaches used in previous methods is pixel-wise direction voting of object keypoints.
In this approach, every pixel votes for the direction to the keypoints and the intersections of the voted directions yield keypoint candidates whose voting scores are then evaluated based on the ratio of voter inliers.
The candidate with the highest scores is decided as the keypoint estimate, and the object pose can be recovered from all keypoint estimates by solving PnP.
The intuition behind this method is that given an object, the location of the invisible keypoint can be inferred from the visible parts \cite{posecnn,pvnet,hybridpose}.
Though this method is known to be robust to occlusion, it is built upon the assumption that the voting directions can reliably yield keypoint candidates regardless of the geometry of the objects.
This assumption is not true.
A counterexample is illustrated in Section \ref{sec:toy:exp} and Figure \ref{fig:toy}.

To address this problem, we proposed a novel framework for estimating 6D pose of 3D objects from RGB input.
The core of our method is a novel representation for object keypoint named \textbf{Keypoint Distance Field (KDF)}.
Inspired by recent works \cite{keypose,pvnet}, we first predict KDF to localize 2D keypoints through voting, then compute object 6D pose by solving a PnP problem.
Figure \ref{fig:overall:arch} illustrates our framework.
In this section, we will introduce the representation of KDF and describe the corresponding voting scheme.
Then we present the implementation details of our approach.

\subsection{Representation of Keypoint Distance Field}
\label{sec:kdf:vote}

Keypoint Distance Field (KDF) is specifically defined for each 2D projected keypoint.
Suppose that the height and width of the input RGB image are $H$ and $W$ respectively and that the object has $K$ keypoints defined. 
KDF for the $k$-th projected keypoint $(u_k, v_k)$ is a two-dimensional array $\mathbf{D} \in \mathbb{R}^{H\times W}$ of the same size as the input image.
The element of the KDF located at $(u,v)$ stores the 2D Euclidean distance between the element and $(u_k, v_k)$
\begin{equation}
    \mathbf{D}_{u,v} = \big|\big| [u,v] - [u_k, v_k] \big|\big|_2
\end{equation}
Note that the KDF can still be defined even when $(u_k, v_k)$ is outside of the image.
We use a fully convolutional neural network to regress $\mathbf{D}$.
Inspired by previous work on 2D object detection \cite{faster:rcnn}, we adopt the following parameterization for regression
\begin{equation}
    t_\mathbf{D} = \log(\mathbf{D}/r)    
\end{equation}
where $r \in \mathbb{R}^+$ is a hyperparameter.
The value of $r$ is chosen to be close to the geometric mean of maximum and minimum possible distances so that the lower and upper bounds of parameterized distance are symmetric about zero, which is easier for the neural network to regress.
For example, with a $256\times 256$ image input, $r$ can be set to be $16$.

\textbf{Loss Function. }
To predict a set of continuous values, we can either regress the values directly or convert it to classification problems with multiple discretized values.
Since the range of possible distance values is large even after parameterization, we choose to use direct regression for KDF prediction to avoid large discretization error.
We use a standard soft $L_1$ loss function for parameterized distance 
\begin{equation}\label{eq:kdf:loss}
    \mathcal{L}(\mathbf{D}, \mathbf{D}^*) = 
    \frac{1}{H\times W}
    \begin{cases}
        \frac{1}{2e}(t_{\mathbf{D}^*} - t_\mathbf{D})^2, & |t_{\mathbf{D}^*} - t_\mathbf{D}| < e \\
        |t_{\mathbf{D}^*} - t_\mathbf{D}| - \frac{e}{2}, & |t_{\mathbf{D}^*} - t_\mathbf{D}| \ge e
    \end{cases}
\end{equation}
where $\mathbf{D}^*$ is the ground truth KDF and $e$ is a threshold value.
Note the loss function is the mean of all elements.

\textbf{Objects with Symmetry.}
Symmetric objects can cause ambiguity among mutually equivalent rigid transforms.
A group of keypoints placed on the object may thus be indistinguishable.
Inspired by \cite{keypose}, to deal with objects with discrete symmetry, we define keypoints such that each keypoint is part of the symmetry permutation group and then apply a permutation loss during training.
Suppose keypoints $k_1, k_2, \ldots, k_S$ are mutually equivalent and the symmetric permutation of their indices is $\text{Sym}(S) = \{(i_1, i_2, \ldots, i_S )\}$.
Then the KDF regression loss under symmetric permutation is
\begin{equation}\label{eq:symm}
    \mathcal{L}_{\text{symm}} = \min_{s\in \text{Sym}(S)} \mathcal{L}(\mathbf{D}, \mathbf{D}_s^*)
\end{equation}
where $\mathbf{D}_s^*$ is the ground truth KDFs after applied with permutation $s$ on keypoints indices. 

For objects with continuous symmetry such as cylinders, the symmetric permutation group is infinite.
Therefore, Equation \eqref{eq:symm} cannot be used. 
Alternatively, we define keypoints on the rotation axis, and predicting at least two keypoints is sufficient for determining the 6D pose of the object with the ambiguity of the rotation.

\subsection{Distance-based Voting Scheme}

In 2D heatmap-based methods \cite{keypose,6dof:kp}, only the small image area that is close to a keypoint and has high probabilities can have an effect on the final predictions.
Therefore, the prediction is affected by the pixels in that small area and suffers from occlusion.
On the contrary, we apply a RANSAC-based voting scheme to take distance predictions from more pixels into account, so that occluded keypoints or even keypoints outside the image can be robustly handled. 

The first step is to generate hypotheses of keypoint locations through sampling.
For keypoint $k$, we first determine the set of elements $\mathcal{P}=\{ \mathbf{p}_i \}$ on the $k$-th KDF map that will participate in the voting.
From $\mathcal{P}$, we randomly sample three elements $\mathbf{p}_{i_1}, \mathbf{p}_{i_2}, \mathbf{p}_{i_3} \sim \mathcal{P}$.
For every $j\in\{i_1, i_2, i_3\}$, the set of  hypothesized keypoint locations predicted by pixel $\mathbf{p}_j$ is a circle whose center is $\mathbf{p}_j$ and radius equals to $\mathbf{D}_{\mathbf{p}_j}$ --- the predicted KDF value at $\mathbf{p}_j$.
Given a perfect KDF, all three circles are supposed to intersect at one location.
In practice, we find the best possible location(s) agreed by at least two of the three circles using the following procedure.
Each pair of the three circles returns two intersections, but at most one intersection is valid as a keypoint hypothesis.
The third circle is used to decide the valid hypothesis based on which intersection is closer to the third circle.
Mathematically, suppose the two circles predicted by $\mathbf{p}_{i_1}$ and  $\mathbf{p}_{i_2}$ intersects at $\mathbf{h}_{i_1}$ and $\mathbf{h}_{i_2}$
which can be obtained by jointly solving the following two quadratic equations
\begin{equation}
    ||\mathbf{h} - \mathbf{p}_{i_1}||_2 = \mathbf{D}_{\mathbf{p}_{i_1}},
    ||\mathbf{h} - \mathbf{p}_{i_2}||_2 = \mathbf{D}_{\mathbf{p}_{i_2}}
\end{equation}
Then the valid hypothesis generated by $\mathbf{p}_{i_1}$ and  $\mathbf{p}_{i_2}$ is given by
\begin{equation}
    \mathbf{h}_{i_1,i_2} = \underset{\mathbf{h} \in \{ \mathbf{h}_{i_1}, \mathbf{h}_{i_2}\}}{\arg\min} \Big| || \mathbf{h} - \mathbf{p}_{i_3} ||_2 - \mathbf{D}_{\mathbf{p}_{i_3}} \Big| 
\end{equation}
The other two valid hypotheses $\mathbf{h}_{i_2,i_3}$ and $\mathbf{h}_{i_3,i_1}$ can be obtained similarly.
In total, there are three valid hypotheses generated.
The above sampling and hypothesis generation is repeated for $N$ times to generate $3N$ hypotheses $\mathcal{H} = \{\mathbf{h}_l \mid l=1,2,\ldots, 3N \}$ for the $k$-th keypoint.
Next, all elements from $\mathcal{P}$ vote for these hypotheses.
The voting score $v_l$ of $\mathbf{h}_l$ is the number of elements whose distance prediction error is within the threshold $\theta$
\begin{equation}\label{eq:vote:score}
    v_l = \sum_{\mathbf{p}\in \mathcal{P}} \mathbb{I} \Big( \Big| || \mathbf{h}_l - \mathbf{p} ||_2 - \mathbf{D}_\mathbf{p} \Big| < \theta \Big)
\end{equation}
where $\mathbb{I}(\cdot)$ is the indicator function.
Then the $k$-th keypoint is determined as the hypothesis with the highest voting score $\mathbf{h}_{l_{\max}}$, where $l_{\max}=\underset{l}{\arg\max} \{v_l\}$.
In this way, we find the location agreed by most of the voters within an error range.
The above process is repeated for all KDFs to predict all keypoints.

\begin{figure}[t]
\centering
\small
\subfloat[]{%
    \includegraphics[height=0.215\linewidth]{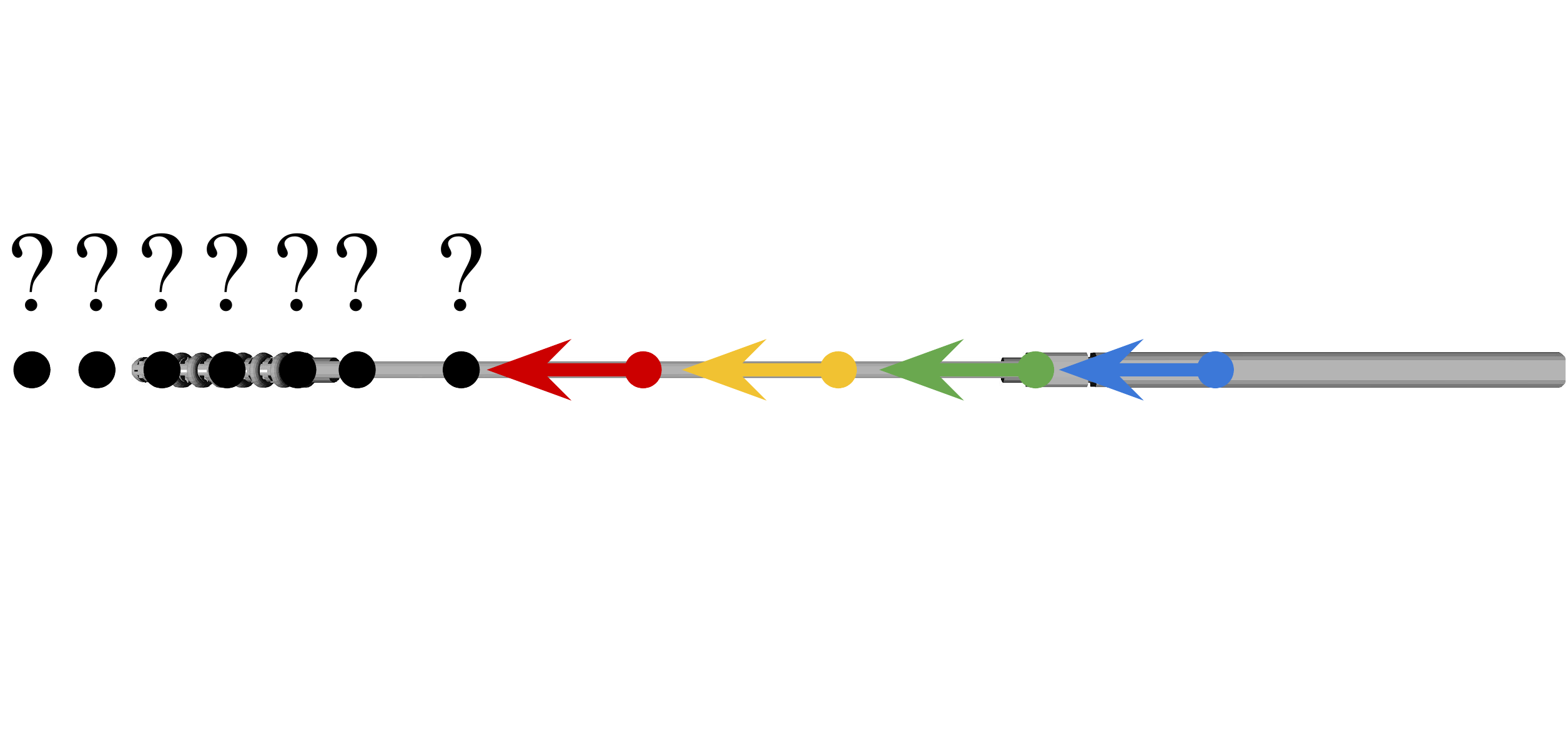}
    \label{fig:dir:vote}
}
\hspace{3ex}
\subfloat[]{%
    \includegraphics[height=0.215\linewidth]{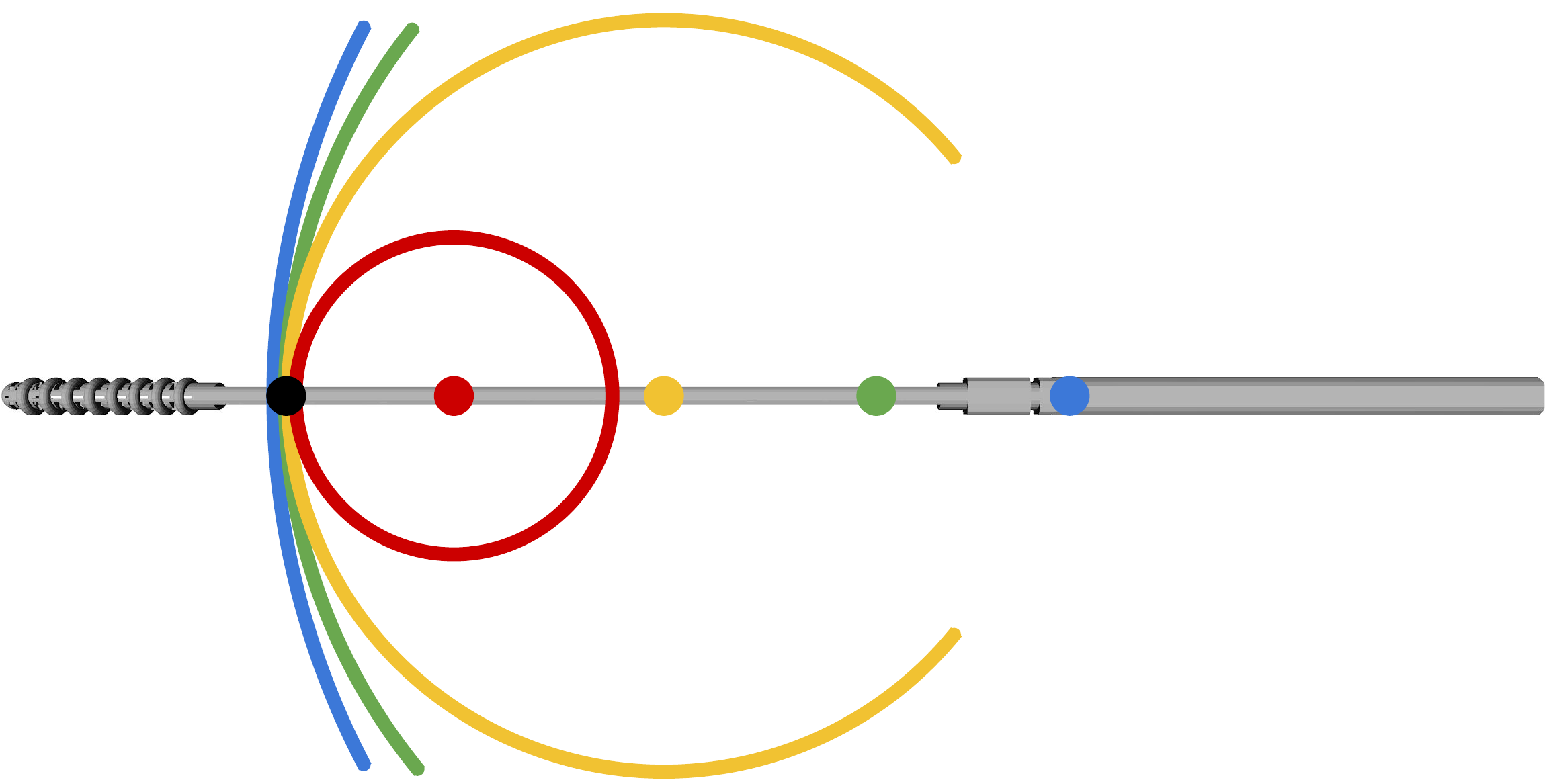}
    \label{fig:dist:vote}
}
\vspace{-1ex}
\caption{Voting scheme comparison. (a) \textbf{Direction-based voting} cannot reliably localize a point when the angles between directions are too small. (b) \textbf{Distance-based voting} can still accurately localize a point in this situation.
}
\label{fig:voting:compare}
\end{figure}

\subsection{Overall Framework and Implementation}
\label{sec:overall:framework}

The KDF prediction CNN is instantiated by a segmentation network with ResNet \cite{resnet} as encoder backbone and additional up-convolution layers and skip connections as the decoder.
The ResNet backbone is initialized with an ImageNet pre-trained model and then finetuned on 6D pose estimation datasets.
During the training of KDF regression network, we randomly generate bounding box crops around the object to introduce more variations.
Random photometric data augmentation is used during training.
The overall architecture is illustrated in Figure \ref{fig:overall:arch}.

Though loss in Equation \eqref{eq:kdf:loss} is applied to all elements, to reduce regression error, one can choose to apply the loss only on elements within a certain keypoint distance during training if the image size is too large.
In this case, the predicted KDF and the ground truth KDF will be different at large keypoint distances during inference,
and a rough initial estimate of the object location from detection or segmentation is needed and $\mathcal{P}$ in Equation \eqref{eq:vote:score} only includes elements within a certain keypoint distance.
We will show in Section \ref{sec:exp:viz} that such training loss strategy does not affect the voting of the 2D keypoints. 
In practice, within a rough initial range of keypoint 2D location, we randomly sample 4,096 pixels as the set of voters $\mathcal{P}$.
Among the voters, we sample 1,024 three tuples of pixels to generate 3,072 keypoint hypotheses whose voting scores are determined by all the sampled voters.
Given the predicted 2D keypoint locations for each object, the 6D pose can be computed by solving a PnP problem.

\begin{table}[t]
\small
\centering
\setlength{\tabcolsep}{4.pt}
\vspace{3ex}
\caption{ \textbf{Toy dataset results.} Evaluation metric is \textbf{2D projection error at 1px threshold}. }
\label{tab:toy}
\begin{tabular}{l|cc}
\hline
method & PVNet \cite{pvnet} & \textbf{KDFNet (ours)} \\
\hline
GT mask & 93.2 & \textbf{95.5} \\
GT mask $+$ keypoint occlusions & 75.1 & \textbf{95.6} \\
\hline
\end{tabular}
\end{table}

\begin{figure}[t]
\centering
\small
\begin{tabular}{@{}c@{}}
  \subfloat[]{\includegraphics[width=0.23\linewidth]{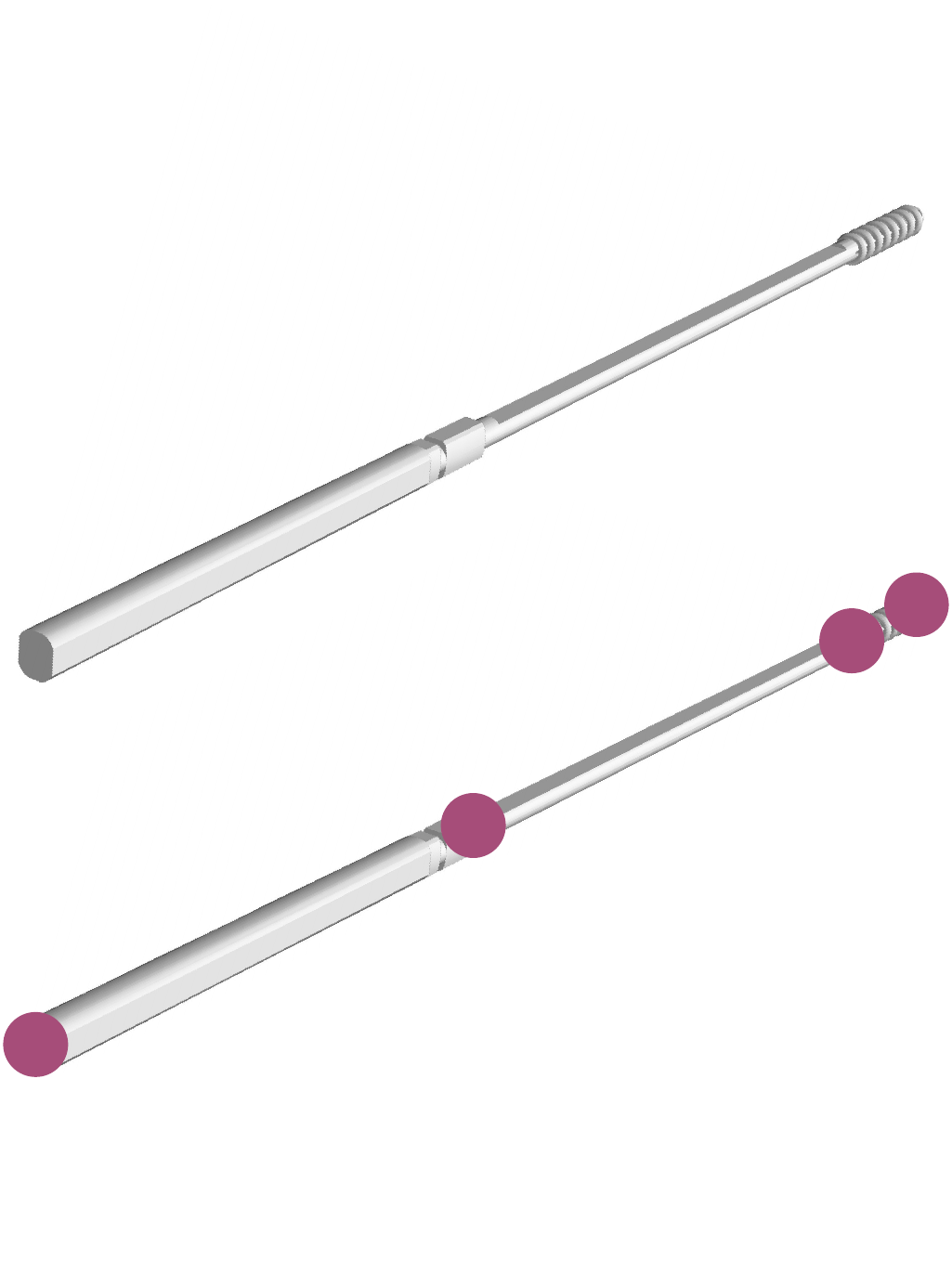}
  \label{fig:toy:obj}
  }
\end{tabular}
\hspace{1ex}
\begin{tabular}{@{}c@{}}
  \subfloat[]{\includegraphics[height=0.45\linewidth]{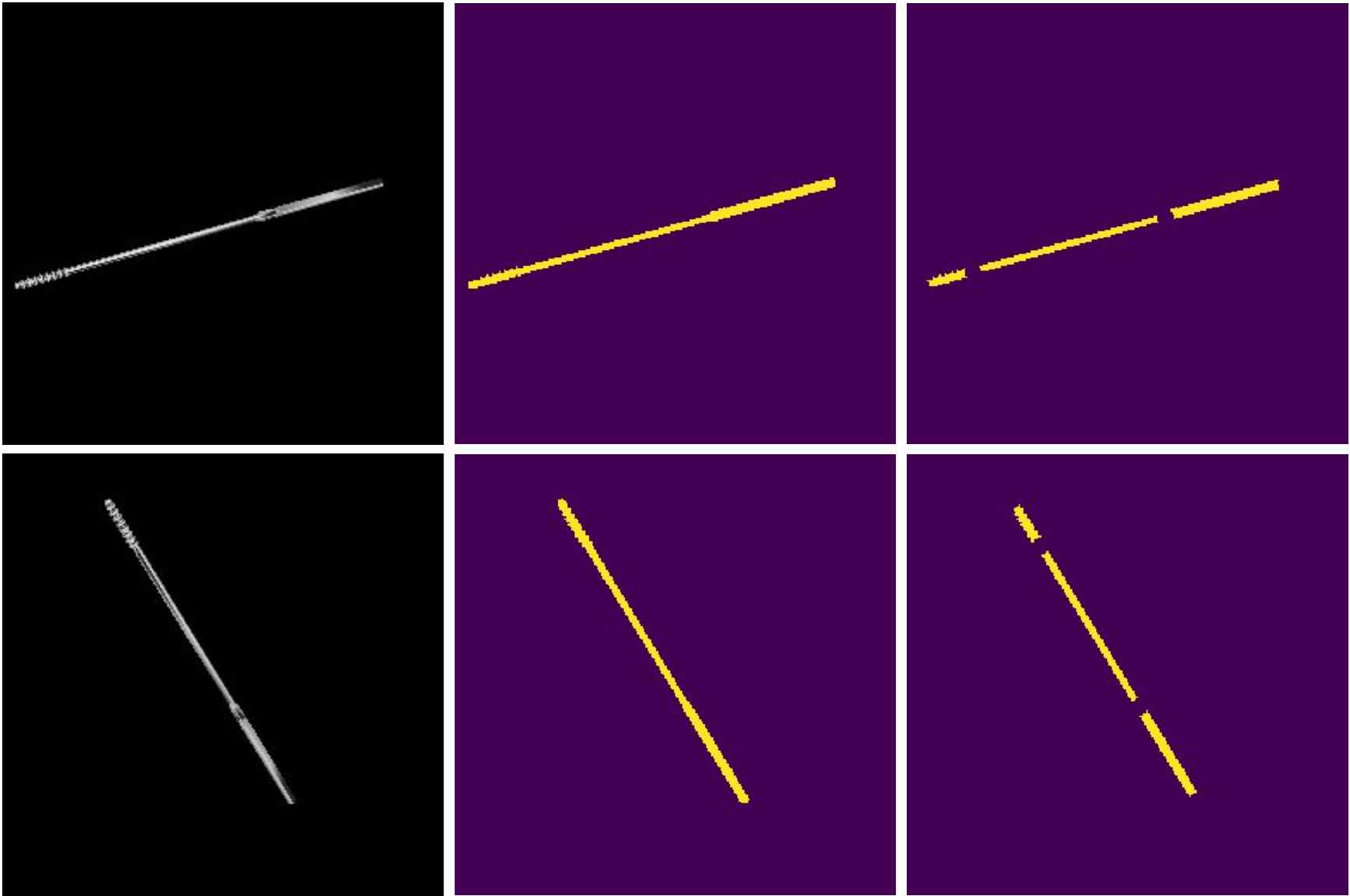}
  \label{fig:toy:data}
  }
\end{tabular}
\caption{\textbf{Toy dataset.} 
(a) The 3D model of the object  used in the toy dataset: a medical swab stick. 
There are four keypoints defined which are shown as purple circles.
(b) Two data samples drawn from the toy dataset. Columns 1-3 are respectively the RGB image input, ground truth object mask, and ground truth object mask with additional occlusions applied around the keypoints.
}
\label{fig:toy}
\end{figure}

\section{Direction-based vs. Distance-based Voting: \\ A Toy Experiment}
\label{sec:toy:exp}

The key difference between our method and previous works \cite{posecnn,pvnet,hybridpose} is the predicted representation used in voting, i.e. keypoint distance vs. keypoint direction.
In this section, we construct a toy synthetic dataset where previous direction-based voting methods fail in localizing the keypoints. 
Through this simple dataset, we show the drawbacks of direction-based representation and the advantage of our proposed KDF representation.

The object we use in the dataset is the 3D mesh model of a medical swab stick for COVID-19 testing, illustrated in Figure \ref{fig:toy}\subref{fig:toy:obj}.
The dataset consists of synthetically rendered images of the swab stick in various 6D poses.
The translation part of the poses $\mathbf{T}$ is the same for every image and the rotation part $\mathbf{R}$ is uniformly randomly sampled.
The backgrounds are all black.
The dataset has 30,000 training and 3,000 validation images. 
The first column of Figure \ref{fig:toy}\subref{fig:toy:data} illustrates two examples of images in our dataset.
Semantic masks are provided for every image.
We define four keypoints along the medical swab stick.

The baseline we compare our KDFNet against is PVNet \cite{pvnet}, a direction-based keypoint method.
To fairly compare, we assume the voters of both PVNet and our KDFNet are the pixels that belong to the same masks.
We test two types of masks: 1) ground truth object mask, and 2) ground truth object mask applied with additional small circular occlusions at keypoint locations.
For evaluation metrics, we measure average 2D projection error which is described in more detail in Section \ref{sec:evaluation:metrics}.
The evaluation threshold is 1 pixel.
The results are illustrated in Table \ref{tab:toy}.
Our model is not affected by occlusions on pixel voters.
However, the performance of PVNet drops significantly.

We provide an explanation as follows.
The geometry of the medical swab stick is long and thin, in which case most of the keypoint voting directions are close to being on the same line except for the small area around the keypoint.
As illustrated in Figure \ref{fig:voting:compare}, intersections cannot be robustly found from the voting directions if the small area near the keypoints is occluded.
On the contrary, our model is based on distance voting and can still reliably locate the keypoints from circle intersection under occlusion.
Through this extremely simple dataset, we show the drawbacks of direction-based keypoint voting methods and the advantage of distance-based voting adopted by our KDFNet.

\section{Experiments}

In this section, we describe the details of our experiment settings in terms of the dataset, implementation details, and evaluation metrics.
We then present the experiment results including ablation studies. 
Additionally, we provide visualizations on KDF and predicted 6D poses of objects.

\subsection{Dataset and Implementation}

\begin{table*}[h]
\setlength{\tabcolsep}{1.2pt}
\centering
\small
\vspace{1ex}
\caption{ \textbf{ADD(-S) accuracy} (left) and \textbf{2D projection error accuracy} (right) results on Occlusion LINEMOD \cite{occlusion:linemod}. }
\label{tab:linemod:results}
\centering
\subfloat{\hspace{0ex}
\begin{tabular}{l|ccccccc}
\hline
methods & 
\multicolumn{1}{c}{\begin{tabular}[c]{@{}c@{}} PoseCNN \\ \cite{posecnn} \end{tabular}} &
\multicolumn{1}{c}{\begin{tabular}[c]{@{}c@{}} Oberweger \\ \cite{oberweger} \end{tabular}} &
\multicolumn{1}{c}{\begin{tabular}[c]{@{}c@{}} Pix2Pose \\ \cite{pix2pose} \end{tabular}} &
\multicolumn{1}{c}{\begin{tabular}[c]{@{}c@{}} PVNet \\ \cite{pvnet} \end{tabular}} &
\multicolumn{1}{c}{\begin{tabular}[c]{@{}c@{}} HybridPose \\ \cite{hybridpose} \end{tabular}} &
\multicolumn{1}{c}{\begin{tabular}[c]{@{}c@{}} \textbf{KDFNet} \\ \textbf{(ours)} \end{tabular}}
\\ 
\hline
\texttt{ape} & 9.6 &  17.6 & \textbf{22.0} & 15.8 & 20.9 & 19.5 \\
\texttt{can} & 45.2 & 53.9 &  44.7 & 63.3 & 75.3 & \textbf{78.4} \\
\texttt{cat} & 0.9 & 3.31 & 22.7 & 16.7 & 24.9 & \textbf{28.2} \\
\texttt{driller} & 41.4 & 45.2  & 44.7 & 65.7 & 70.2 & \textbf{75.1}  \\
\texttt{duck} & 19.6 & 17.2 & 15.0 & 25.2 & 27.9 & \textbf{38.6} \\
\texttt{eggbox} & 22.0 & 25.9 & 25.2 & 50.1 & \textbf{52.4} & 51.2 \\
\texttt{glue} & 38.5 & 39.6 & 32.4 & 49.6 & \textbf{53.8} & 52.1 \\
\texttt{holepuncher} & 22.1 & 21.3 & 49.5 & 12.4 & 54.2 & \textbf{59.0} \\
\hline
average & 24.9 & 30.4 & 32.0 & 40.8 & 47.5 & \textbf{50.3} \\ 
\hline
\end{tabular}
}
\hspace{0.1ex}
\subfloat{\hspace{0ex}
\begin{tabular}{l|cccc}
\hline
methods &
\multicolumn{1}{c}{\begin{tabular}[c]{@{}c@{}} PoseCNN \\ \cite{posecnn} \end{tabular}} &
\multicolumn{1}{c}{\begin{tabular}[c]{@{}c@{}} Oberweger \\ \cite{oberweger} \end{tabular}} &
\multicolumn{1}{c}{\begin{tabular}[c]{@{}c@{}} PVNet \\ \cite{pvnet} \end{tabular}} &
\multicolumn{1}{c}{\begin{tabular}[c]{@{}c@{}} \textbf{KDFNet} \\ \textbf{(ours)} \end{tabular}} \\
\hline
\texttt{ape} & 34.6 & \textbf{69.6} & 69.1 & 66.6 \\
\texttt{can} & 15.1 & 82.6 & 86.1 & \textbf{91.1} \\
\texttt{cat} & 10.4 & 65.1 & 65.1 & \textbf{72.5} \\
\texttt{driller} & 31.8 & 73.8 & 73.1 & \textbf{79.7} \\
\texttt{duck} & 7.4 & 61.4 & 61.4 & \textbf{71.3} \\
\texttt{eggbox} & 1.9 & \textbf{13.1} & 8.4 & 6.1 \\
\texttt{glue} &  13.8 & 54.9 & 55.4 & \textbf{59.6} \\
\texttt{holepuncher} &  23.1 & 66.4 & 69.8 & \textbf{85.3} \\
\hline
average & 17.2 & 60.9 & 61.1 & \textbf{66.5} \\ 
\hline
\end{tabular}
}

\end{table*}

\textbf{Occlusion LINEMOD \cite{occlusion:linemod}} is a standard benchmark for 6D object pose estimation.
It contains videos of desktop objects in a cluttered scene.
It is a subset of the LINEMOD dataset \cite{linemod} that mainly focuses on objects under occlusion.
Together with the annotated images, high-quality 3D scanned models of the objects are also provided.
The test set of Occlusion LINEMOD consists of an image sequence of 1,214 frames, each annotated with the 6D poses of 8 objects from LINEMOD dataset.
The data used to train our model are the same as \cite{pvnet}: real images from LINEMOD and synthetically rendered images using the scanned 3D object models.
We adopt the same object keypoint set as \cite{pvnet}, which are generated by Farthest Point Sampling (FPS) of the object 3D point set. 

\textbf{TOD Dataset \cite{keypose}} is a dataset for keypoint estimation of transparent objects.
It consists of 48,000 stereo images from 600 stereo videos of 15 transparent objects placed on simple textured tabletops.
Every stereo image is annotated with 3D keypoints.
Together with the images, high-quality 3D scanned models of the objects are also provided.
In TOD dataset, there are 2,880 training images and 320 test images for each object.
Since the objects are transparent, we did not render additional synthetic images during training but adopt geometric and photometric data augmentations during training.
Though TOD dataset was originally created for keypoint estimation only, it can still be used to train and evaluate the 6D poses of the objects.
To this end, we select a subset of the objects in TOD that have at least three keypoints defined, i.e. the 7 \texttt{mug} categories, so that there are enough keypoints for recovering 6D poses.

\subsection{Evaluation Metrics} \label{sec:evaluation:metrics}
\textbf{ADD(-S) Metrics.}
We use ADD \cite{linemod} and ADD-S \cite{posecnn} in our evaluation.
When computing ADD distance, we transform the model point set by the predicted and the ground truth poses respectively, and compute the mean 3D Euclidean distance between the two point sets.
Given an object with 3D model point set of $\mathcal{M} = \{ \mathbf{x}_i \in \mathbb{R}^3 \mid i=1,2,\ldots, M \}$, the ADD distance is calculated as
\begin{equation}\label{eq:add}
   \text{ADD} = \frac{1}{|\mathcal{M}|} \sum_{\mathbf{x}\in\mathcal{M}} || (\mathbf{R}\mathbf{x} + \mathbf{T}) - (\mathbf{R}^*\mathbf{x} + \mathbf{T}^*) || 
\end{equation}
where $\mathbf{R}^*$ and $\mathbf{R}$ are the ground truth and estimated rotation, and $\mathbf{T}^*$ and $\mathbf{T}$ are the ground truth and estimated
translation.
For symmetric objects, ADD-S \cite{posecnn} is used instead.
When computing ADD-S distance, the 3D distances are calculated between the closest points
\begin{equation}\label{eq:add:s}
   \text{ADD-S} = \frac{1}{|\mathcal{M}|} \sum_{\mathbf{x}_1\in\mathcal{M}} \min_{\mathbf{x}_2\in\mathcal{M}} || (\mathbf{R}\mathbf{x}_1 + \mathbf{T}) - (\mathbf{R}^*\mathbf{x}_2 + \mathbf{T}^*) || 
\end{equation}

We use the following two evaluation metrics. 
(1) \textbf{ADD(-S) accuracy}: ADD(-S) accuracy measures the proportion of correct pose predictions. A pose prediction is considered correct if the ADD(-S) distance is less than the threshold of 10\% of the model's diameter. 
(2) \textbf{ADD(-S) AUC}: the area under ADD(-S) accuracy-threshold curve where the maximum threshold is set to 10cm.
To compare with previous works, we use ADD(-S) accuracy on Occlusion LINEMOD datasets, and ADD accuracy and AUC on TOD dataset.

\textbf{2D Projection Metric.}
When computing 2D projection error, we transform the model point set by the predicted and the ground truth poses respectively, and compute the mean 2D distance between the image projections of model points.
Given the camera projection function $\Pi: \mathbb{R}^3 \rightarrow \mathbb{R}^2$, the 2D projection error is calculated as
\begin{equation}\label{eq:2d:proj}
   \text{2D-Proj} = \frac{1}{|\mathcal{M}|} \sum_{\mathbf{x}\in\mathcal{M}} || \Pi(\mathbf{R}\mathbf{x} + \mathbf{T}) - \Pi(\mathbf{R}^*\mathbf{x} + \mathbf{T}^*) || 
\end{equation}

A pose prediction is considered as correct if the distance is less than the threshold of 5 pixels.
We use the \textbf{2D projection accuracy} to evaluate Occlusion LINEMOD dataset.

\subsection{Experiment Results and Comparison}
We train the KDFNet model on the LINEMOD dataset and rendered synthetic images.
We evaluate and compare our model against previous baselines \cite{posecnn,oberweger,pix2pose,pvnet,hybridpose} on Occlusion LINEMOD dataset.
The results are illustrated in Table \ref{tab:linemod:results}.
Among these baselines, the most relevant is PVNet \cite{pvnet}, a direction-based keypoint voting method which was also compared against in the toy experiment in Section \ref{sec:toy:exp}.
Our method achieves the best average ADD(-S) accuracy of 50.3\% among all baselines while being the best on 5 of the 8 objects.
In terms of 2D projection accuracy, our model also achieves the best among all baselines with an average of 66.5\% while being the best on 6 of the 8 objects.
In particular, our method outperforms PVNet \cite{pvnet} by a margin of 9.5\% in ADD(-S) accuracy and 5.6\% in 2D projection accuracy, while also outperforming previous state-of-the-art HybridPose \cite{hybridpose} by 2.8\% in ADD(-S) accuracy.

Additionally, we train our model on TOD dataset \cite{keypose} to predict the object keypoints in both stereo images and compare it against KeyPose \cite{keypose}.
KeyPose predicts object keypoints using heatmaps in both stereo images and the 6D object poses are computed by keypoint triangulation from stereo and pose fitting by solve an Orthogonal Procrustes problem.
To evaluate KDFNet on TOD dataset, we follow the same procedure in \cite{keypose} to recover 6D object pose from the predicted 2D keypoints in both stereo images.
The results are illustrated in Table \ref{tab:tod}. 
Our method achieves state-of-the-art performance and surpasses \cite{keypose}.
Note that our leading margin is not as significant as on Occlusion LINEMOD dataset. 
An explanation is that TOD dataset did not include occlusion, therefore the proposed KDFNet cannot fully show its ability in dealing with occlusion compared to a heatmap-based method.

\begin{table}[t]
\small
\centering
\setlength{\tabcolsep}{3.3pt}
\caption{ Ablation Study on the number of keypoints $K$. Evaluation metrics are \textbf{ADD(-S) accuracy} and \textbf{2D projection}. }
\label{tab:ablation:num:kp}
\begin{tabular}{l|cccc|cccc}
\hline
metrics & \multicolumn{4}{c|}{ADD(-S) accuracy} & \multicolumn{4}{c}{2D projection} \\
\hline
\# of kps $K$ & 8 & 12 & 16 & 20 & 8 & 12 & 16 & 20 \\
\hline
\texttt{ape} & 18.5 & 19.0 & 19.3 & \textbf{19.5} & \textbf{67.3} & 67.0 & 67.1 & 66.6 \\ 	
\texttt{can} & 77.7 & \textbf{80.0} & 79.4 & 78.4 & 90.6 & \textbf{91.4} & 91.2 & 91.1 \\
\texttt{cat} & 26.9 & 27.2 & \textbf{28.3} & 28.2 &  \textbf{74.9} & 73.8 & 73.1 & 72.5 \\
\texttt{driller} & 72.3 & 74.6 & \textbf{75.5} & 75.1 & 77.0 & 78.4  & 79.4 & \textbf{79.7} \\
\texttt{duck} & 36.9 & 37.4 & 38.0 & \textbf{38.7} & \textbf{72.1} & 71.8 & 71.8 & 71.3 \\
\texttt{eggbox} & 50.6 & 50.5 & 51.7 & 51.3 & 6.1 & 5.7 & 5.6 & 6.1 \\
\texttt{glue} & 49.3 & 50.3 & 51.8 & \textbf{52.1} & 57.7 & 59 & 59.3 & \textbf{59.6} \\
\texttt{holepuncher} & 57.3 & 58.5 & 58.8 & \textbf{59.0} & \textbf{85.5} & 85.4 & \textbf{85.5} & 85.3 \\ \hline
average  & 48.7 & 49.7 & \textbf{50.3} & \textbf{50.3} & 66.4 & 66.5 & \textbf{66.6} & 66.5 \\
\hline
\end{tabular}
\end{table}

\begin{table}[t]
\small
\centering
\setlength{\tabcolsep}{3.3pt}
\caption{ Ablation Study on the threshold value. Evaluation metrics are \textbf{ADD(-S) accuracy} and \textbf{2D projection}. }
\label{tab:ablation:threshold}
\begin{tabular}{l|cccc|cccc}
\hline
metrics & \multicolumn{4}{c|}{ADD(-S) accuracy} & \multicolumn{4}{c}{2D projection} \\
\hline
threshold $\theta$ & 0.2 & 0.4 & 0.8 & 1.6 & 0.2 & 0.4 & 0.8 & 1.6 \\
\hline
\texttt{ape} & 18.6 & 19.5 & 20.0 & \textbf{20.2} & 66.6 & 66.6 & \textbf{66.9} & 66.6 \\
\texttt{can} & \textbf{79.0} & 78.4 & 77.1 & 77.2 & 91.1 &  91.1 & 91.0 & \textbf{91.3} \\
\texttt{cat} & \textbf{28.6} & 28.2 & 26.4 & 25.3 & 72.6 & 72.5 & \textbf{72.9} & \textbf{72.9} \\
\texttt{driller} & \textbf{75.9} & 75.1 & \textbf{75.9} & 75.0 & \textbf{80.2} & 79.7 & 79.5 & 79.0 \\
\texttt{duck} & 37.7 & 38.0 & 38.7 & \textbf{39.4} & 71.6 & 71.3 & \textbf{72.1} & 72.0 \\
\texttt{eggbox} & 49.5 & 51.3 & 52.5 & \textbf{58.0} & \textbf{6.3} & 6.1 & 6.1 & 6.2  \\ 	
\texttt{glue} & 51.4 & \textbf{52.1} & 50.8 & 50.1 & \textbf{59.6} & \textbf{59.6} & 59.2 & 58.8 \\
\texttt{holepuncher} & \textbf{59.1} & 59.0 & 58.5 & 57.1 & \textbf{85.3} & \textbf{85.3} & 84.9 & \textbf{85.3}  \\ \hline
average & 50.0 & \textbf{50.3} & 50.1 & 50.1 & \textbf{66.6} & 66.5 & \textbf{66.6} & 66.5 \\
\hline
\end{tabular}
\end{table}

\begin{table}[t]
\small
\centering
\setlength{\tabcolsep}{2.95 pt}
\caption{ Ablation Study on the number of keypoint hypotheses $3N$. Evaluation metrics are \textbf{ADD(-S) accuracy} and \textbf{2D projection}. }
\label{tab:ablation:num:hypothesis}
\begin{tabular}{l|cccc|cccc}
\hline
metrics & \multicolumn{4}{c|}{ADD(-S) accuracy} & \multicolumn{4}{c}{2D projection} \\
\hline
\# of hypotheses & 48 & 192 & 768 & 3072 & 48 & 192 & 768 & 3072 \\
\hline
\texttt{ape} & 20.0 & 19.9 & 19.3 & 19.5 & 66.9 & 66.8 & 66.7 & 66.6
 \\ 	
\texttt{can} & 78.3 & 78.8 & 78.7 & 78.4 & 91.3 & 91.2 & 91.1 & 91.1
 \\
\texttt{cat} & 28.1 & 28.4 & 28.3 & 28.2 & 72.7 & 72.3 & 72.5 & 72.5 \\
\texttt{driller} & 75.2 & 75.5 & 75.7 & 75.1 & 79.9 & 79.1 & 79.5 & 79.7 \\
\texttt{duck} & 38.9 & 38.5 & 38.4 & 38.7 & 71.7 & 71.5 & 71.6 & 71.3 \\
\texttt{eggbox} & 50.2 & 51.0 & 51.2 & 51.3 & 6.1 & 5.9 & 5.9 & 6.1 \\
\texttt{glue} & 51.1 & 51.2 & 51.6 & 52.1 & 59.6 & 59.8 & 60.4 & 59.6 \\
\texttt{holepuncher} & 59.2 & 58.5 & 59.0 & 59.0 & 85.2 & 85.3 & 85.4 & 85.3 \\ \hline
average  & 50.1 & 50.2 & 50.3 & 50.3 & 66.7 & 66.5 & 66.7 & 66.5 \\
\hline
\end{tabular}
\end{table}

\subsection{Ablation Studies}

We conduct ablation studies on Occlusion LINEMOD dataset in the following three aspects to help and validate the design choices of our architecture.

\textbf{Number of Keypoints $\bm{K}$} defined for the object.
Intuitively, increase the number of keypoints will provide more information for the 2D-3D correspondences and yield better 6D pose estimates. 
In the ablation study, the value of $K$ varies from 8 to 20.
The ablation results are illustrated in Table \ref{tab:ablation:num:kp}.
For most objects, the performance generally increases as the number of keypoints increases.
The performance saturates at $K=16$. which means localizing additional keypoints does not further improve the estimation of 2D-3D correspondence.

\textbf{Pixel Threshold Value $\bm{\theta}$} in Equation \eqref{eq:vote:score}.
We are interested in whether the pose estimation results are sensitive to the choice of $\theta$.
The ablation results are illustrated in Table \ref{tab:ablation:threshold}.
Though the threshold value varies in a wide range from 0.2 to 1.6, both metrics of ADD(-S) and 2D projection accuracy stay almost the same.
This means the choice of the hyperparameter $\theta$ does not exert a significant effect on the performance as long as it is within a reasonable range, which shows the stability with respect to $\theta$ of our model during inference.

\textbf{Number of Keypoint Hypothesis $\bm{3N}$}.
We are interested in the number of keypoint hypotheses that are sufficient for finding good keypoint estimates.
The ablation results are illustrated in Table \ref{tab:ablation:num:hypothesis}.
Starting at $3N=48$, both metrics stay almost the same.
This means with the good regression of KDF, a small number of keypoint hypotheses can cover keypoints that result in near-optimal predictions.

\subsection{Visualization}\label{sec:exp:viz}

We visualize the KDF, voted 2D locations of the keypoints, and the estimated 6D poses on Occlusion LINEMOD dataset in Figure \ref{fig:viz:map}.
The KDFs are visualized as a heatmap superimposed on the RGB images.
Our framework can accurately localize keypoints even under occlusion.
The predicted KDF and the ground truth KDF are different at large keypoint distances because we followed the training loss strategy in \ref{sec:overall:framework} and only train the model on elements that are within the keypoint distance of 64 pixels.
As illustrated in Figure \ref{fig:viz:map}\subref{fig:viz:pred:1}-\subref{fig:viz:pred:2}, such difference does not affect the voting process and our framework can still localize keypoints on the image and estimate 6D poses accurately.

\subsection{Run Time Analysis.}
We test KDFNet on a machine with an Intel i7-6850K 3.7GHz CPU, a GTX
1080 Ti GPU and Tensorflow 1.15.
Given the input image of resolution of $256\times 256$ and 192 keypoint hypotheses, the network inference and voting take 116ms and 24ms respectively,
resulting in KDFNet running at 7 frames/sec.

\section{Conclusions}

In this work, we propose a novel method named KDFNet for 6D pose estimation from RGB images.
Our method is based on the novel representation of Keypoint Distance Field (KDF).
We also proposed a distance-based voting scheme to recover the 2D locations of keypoints from predicted distance fields in a RANSAC fashion.
Experiment results show that  KDFNet achieves state-of-the-art performance on Occlusion LINEMOD and TOD dataset.

As future work, we will investigate the extension of the proposed idea of distance field and voting to robotic perception problems in other scenarios or modalities, such as object detection \cite{offboard:detection} in temporal \cite{cpnet} and/or 3D data \cite{meteornet, flownet3d}.

\section{Acknowledgement}
This work is funded in part by JST AIP Acceleration, Grant Number JPMJCR20U1, Japan.

\begin{table}[t]
\small
\centering
\vspace{1ex}
\caption{\textbf{TOD dataset} results.
Evaluation metrics are 
\textbf{ADD accuracy} (\%) and \textbf{AUC} with range of 0 to 10cm.}
\label{tab:tod}
\begin{tabular}{l|c|c|c|c}
\hline
method & \multicolumn{2}{c|}{KeyPose \cite{keypose}} & \multicolumn{2}{c}{\textbf{KDFNet (ours)}} \\ \hline
metrics & accuracy & AUC & accuracy & AUC \\ 
\hline
\texttt{mug\_0} & 70.57 & \textbf{90.06} & \textbf{71.55} & 89.72 \\ 
\texttt{mug\_1} & 48.09 & 81.73 & \textbf{53.43} & \textbf{85.28} \\ 
\texttt{mug\_2} & 67.72 & \textbf{88.95} & \textbf{73.42} &
86.82 \\ 
\texttt{mug\_3} & 72.50 & \textbf{88.59} & \textbf{76.88} & 88.14 \\ 
\texttt{mug\_4} & \textbf{75.00} & 85.69 & 74.69 & \textbf{86.21} \\ 
\texttt{mug\_5} & 91.48 & 91.14 & \textbf{92.43} &
\textbf{91.81} \\ 
\texttt{mug\_6} & 87.66 & \textbf{89.83} & \textbf{87.66} & 89.73 \\
\hline
average & 73.29 & 88.00 &  \textbf{75.72} & \textbf{88.24}  \\ 
\hline
\end{tabular}
\vspace{-1ex}
\end{table}

\newcommand\vizheight{0.141}

\begin{figure*}[t]
\centering
\small
\captionsetup[subfigure]{position=top}

\subfloat[]{%
    \includegraphics[height=\vizheight\linewidth, valign=t]{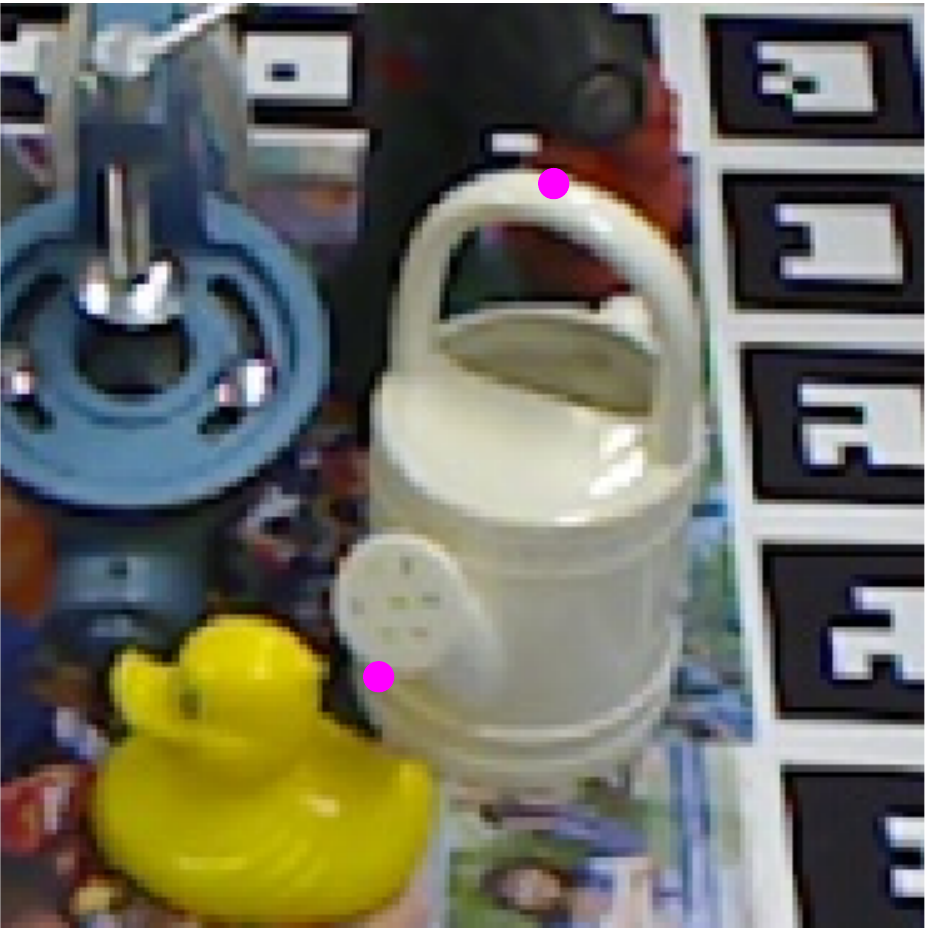}
}
\hspace{0.4ex}
\subfloat[]{%
    \includegraphics[height=\vizheight\linewidth, valign=t]{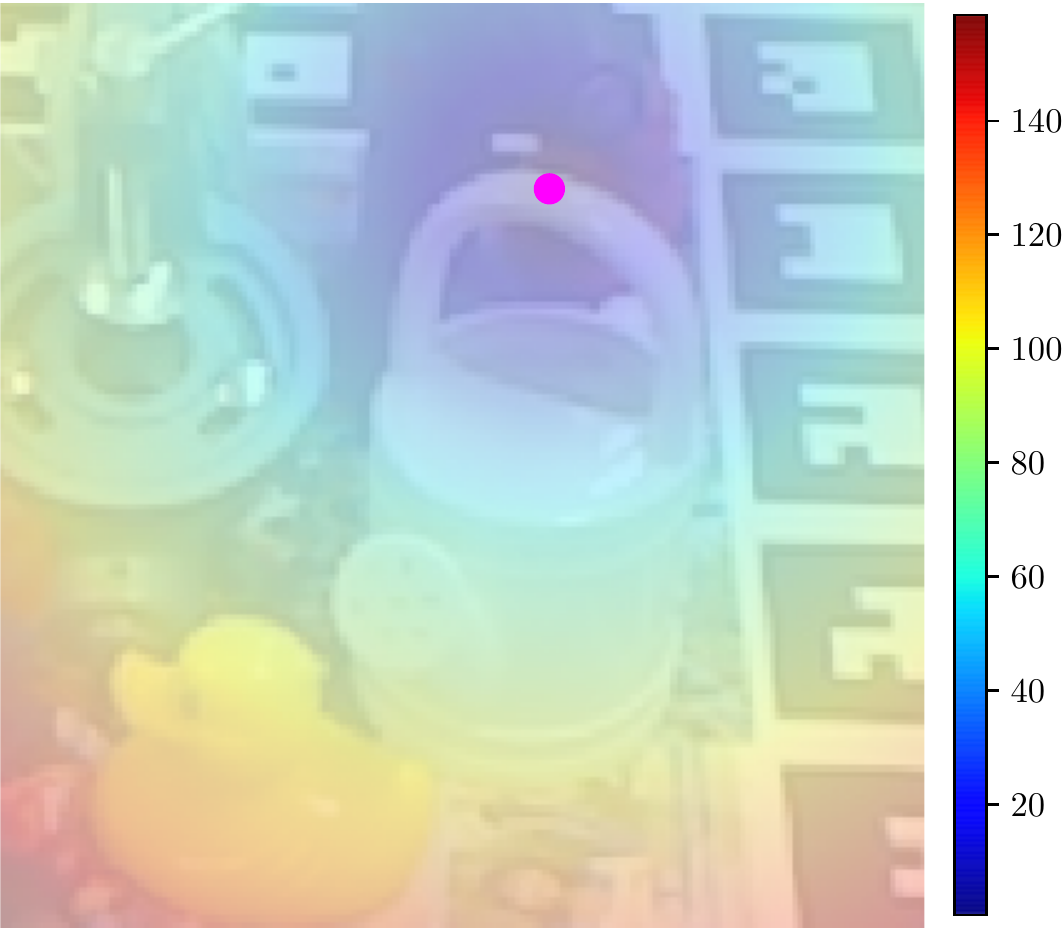}
}
\subfloat[]{%
    \includegraphics[height=\vizheight\linewidth, valign=t]{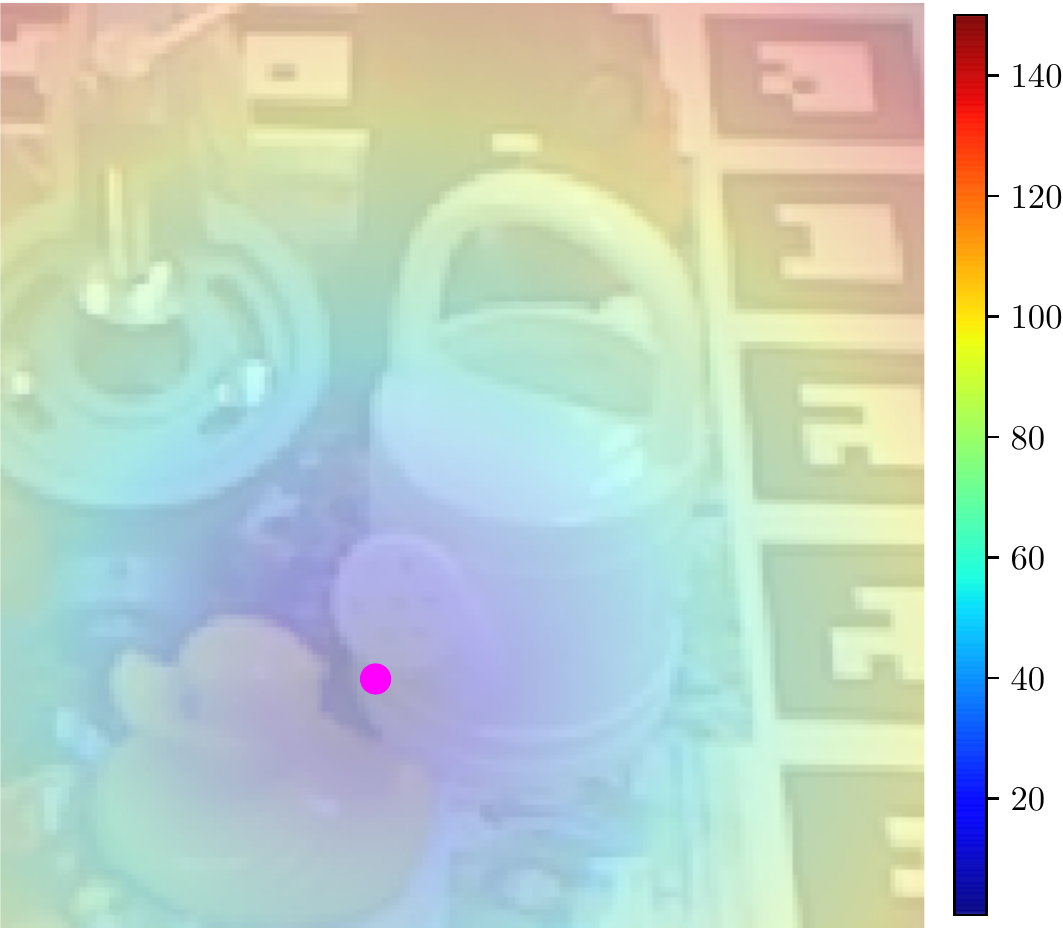}
}
\subfloat[]{%
    \includegraphics[height=\vizheight\linewidth, valign=t]{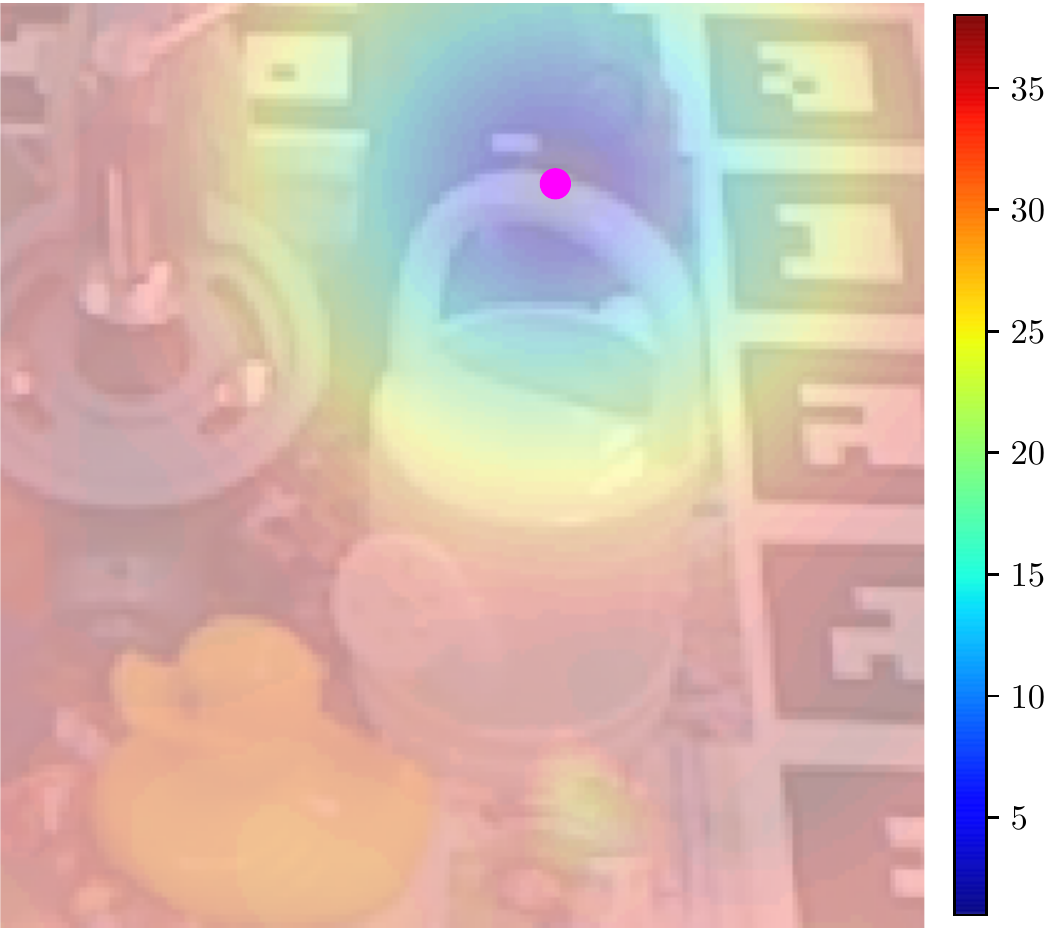}
    \label{fig:viz:pred:1}
}
\subfloat[]{%
    \includegraphics[height=\vizheight\linewidth, valign=t]{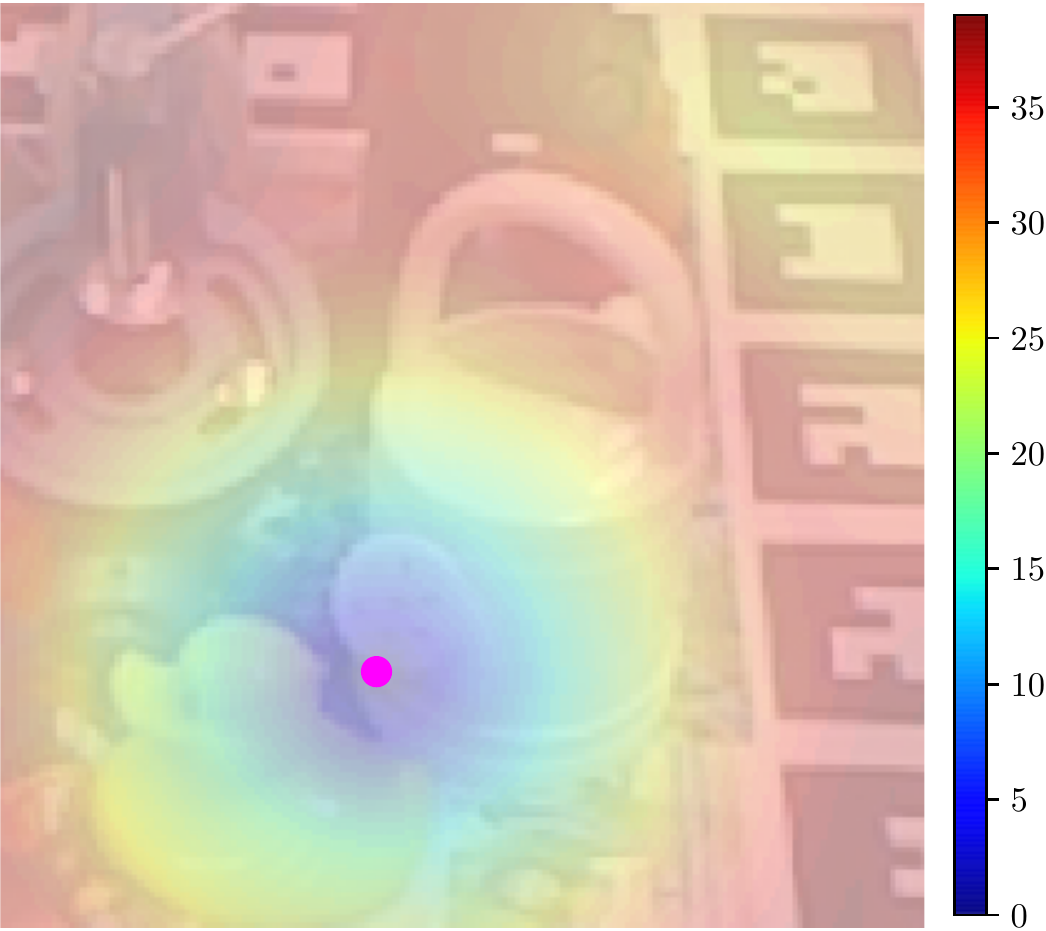}
    \label{fig:viz:pred:2}
}
\subfloat[]{%
    \includegraphics[height=\vizheight\linewidth, valign=t]{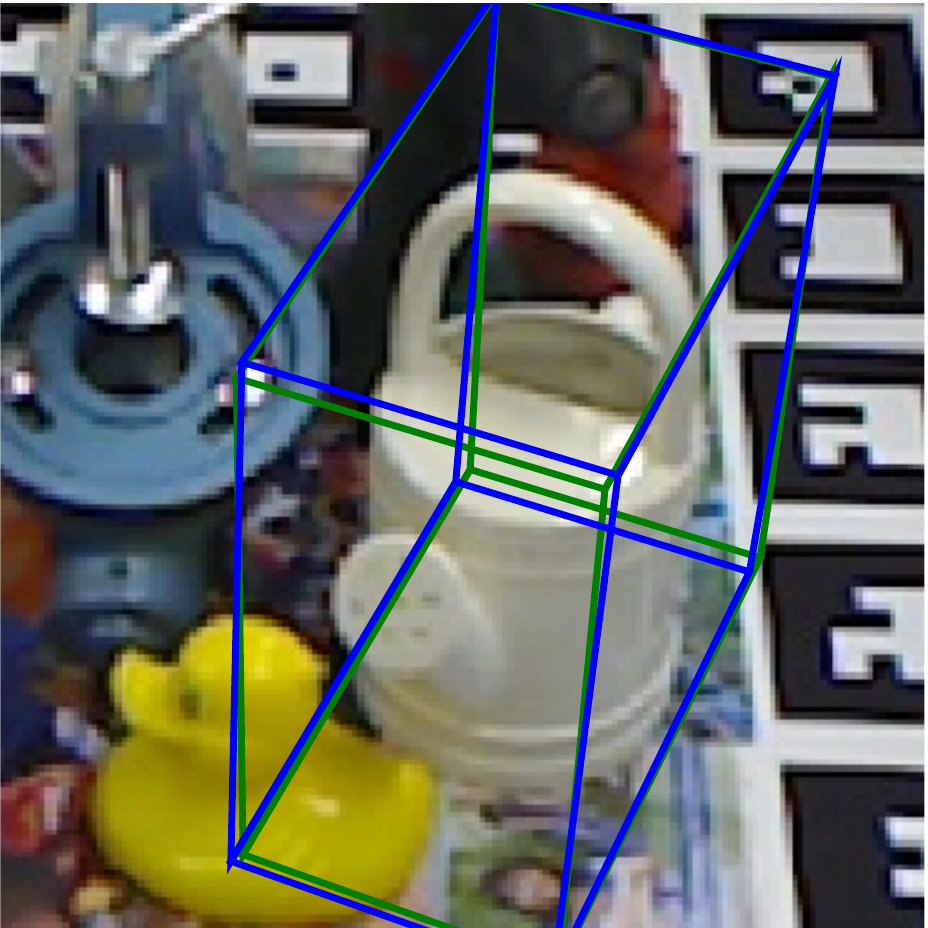}
}
\\
\subfloat{%
    \includegraphics[height=\vizheight\linewidth, valign=t]{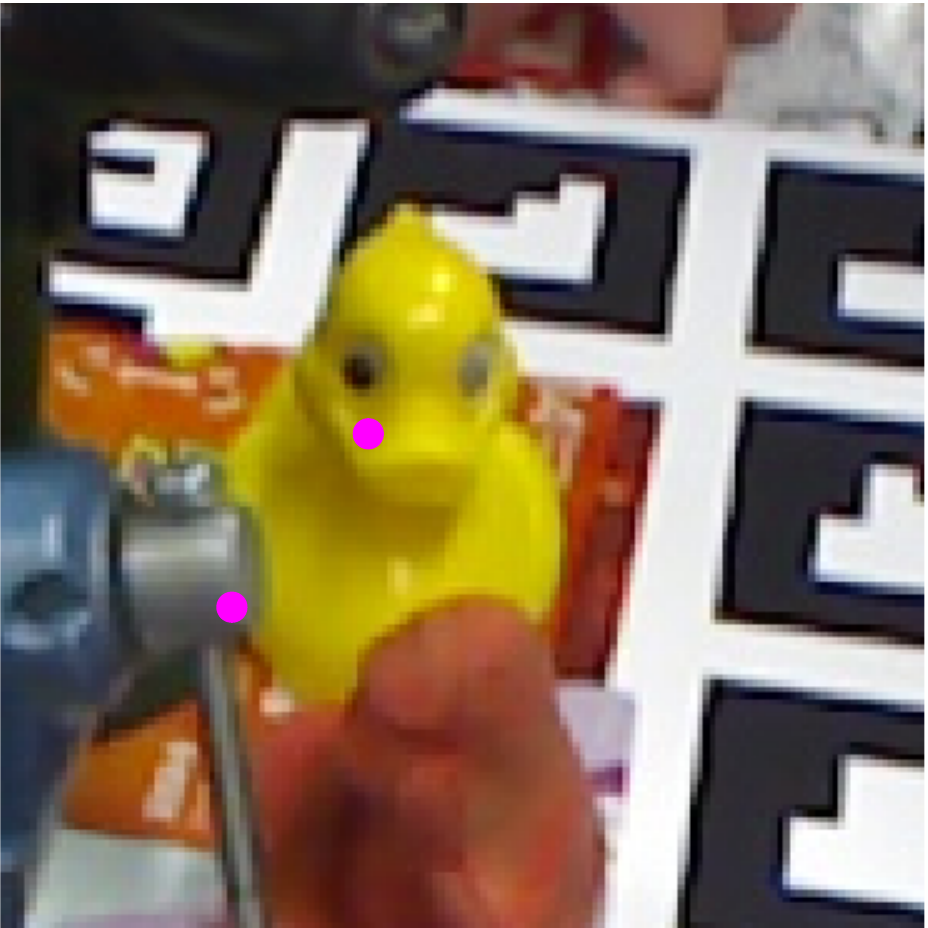}
}
\hspace{0.4ex}
\subfloat{%
    \includegraphics[height=\vizheight\linewidth, valign=t]{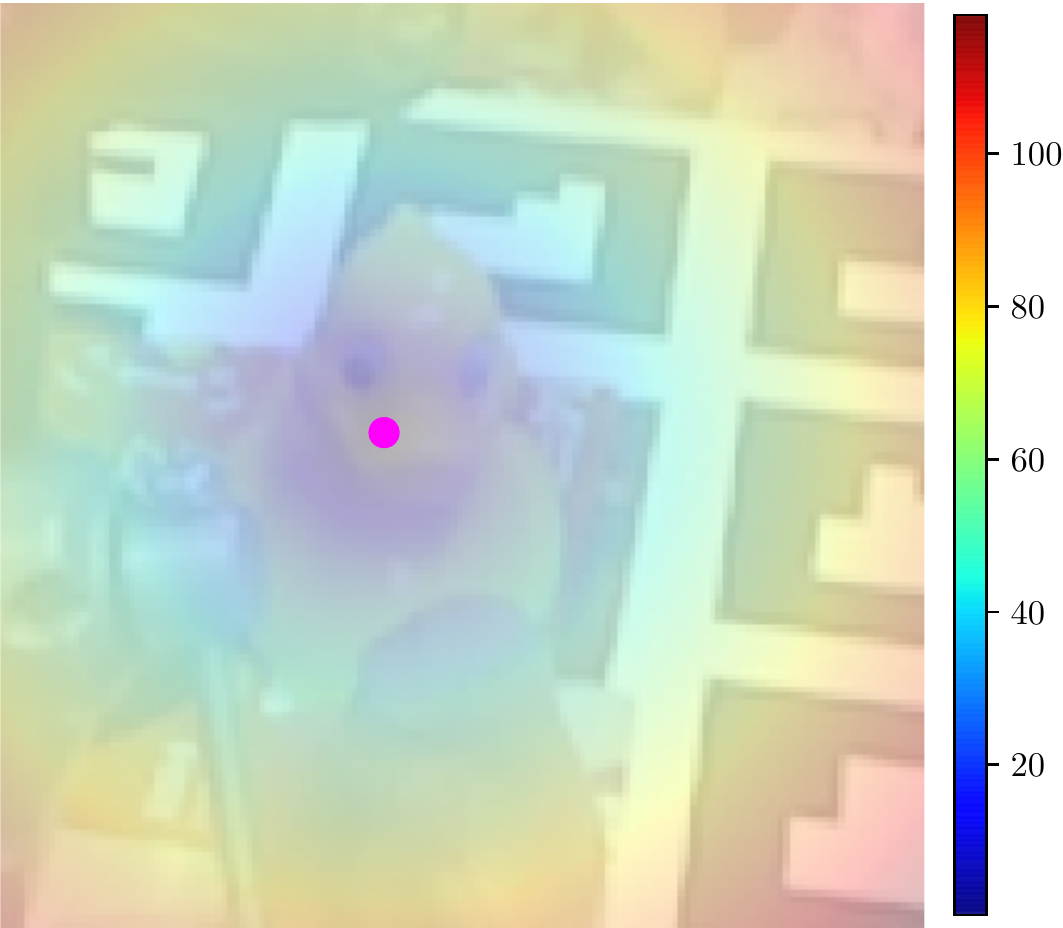}
}
\subfloat{%
    \includegraphics[height=\vizheight\linewidth, valign=t]{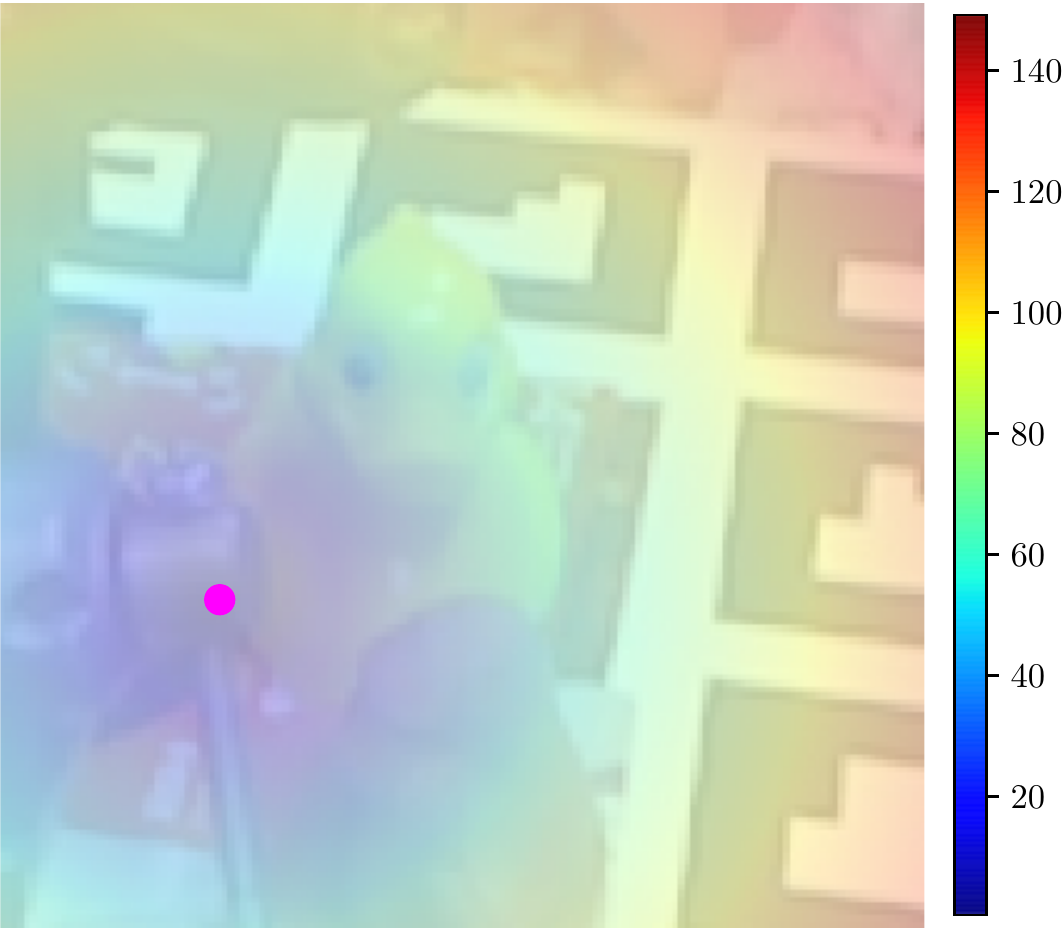}
}
\subfloat{%
    \includegraphics[height=\vizheight\linewidth, valign=t]{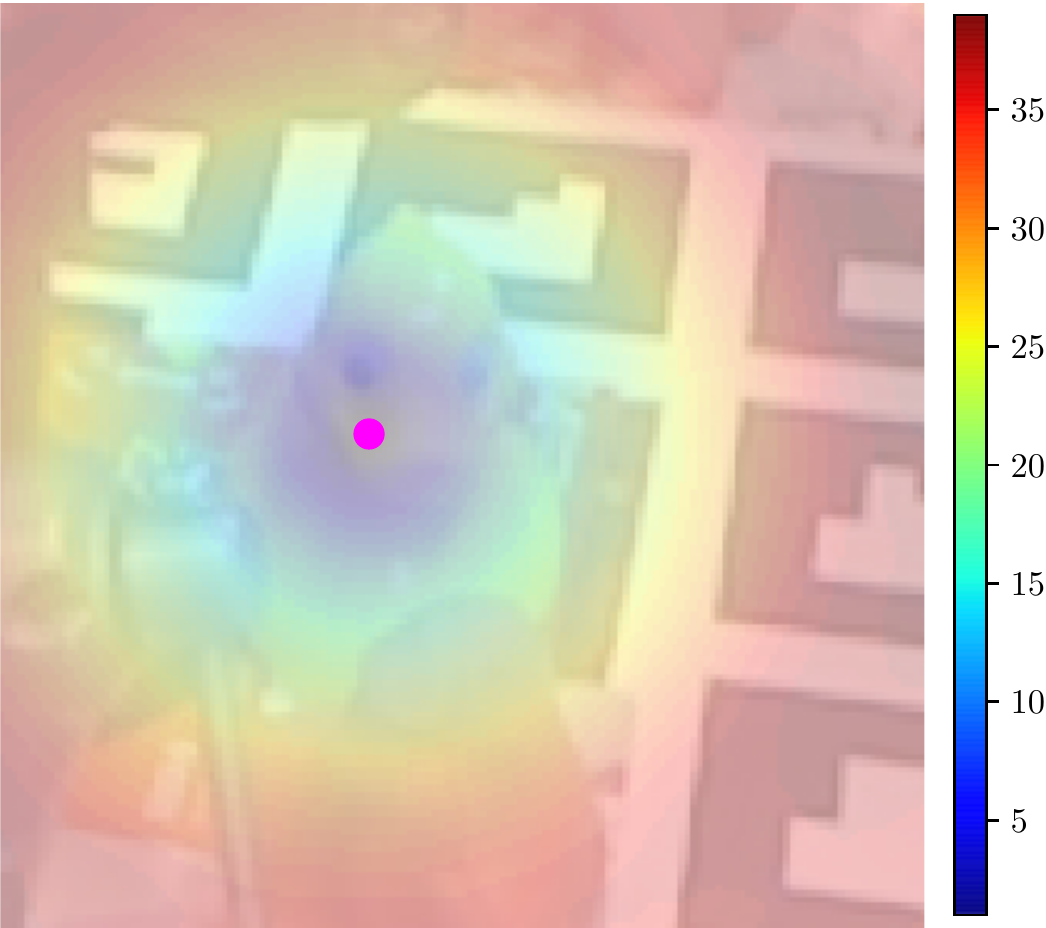}
}
\subfloat{%
    \includegraphics[height=\vizheight\linewidth, valign=t]{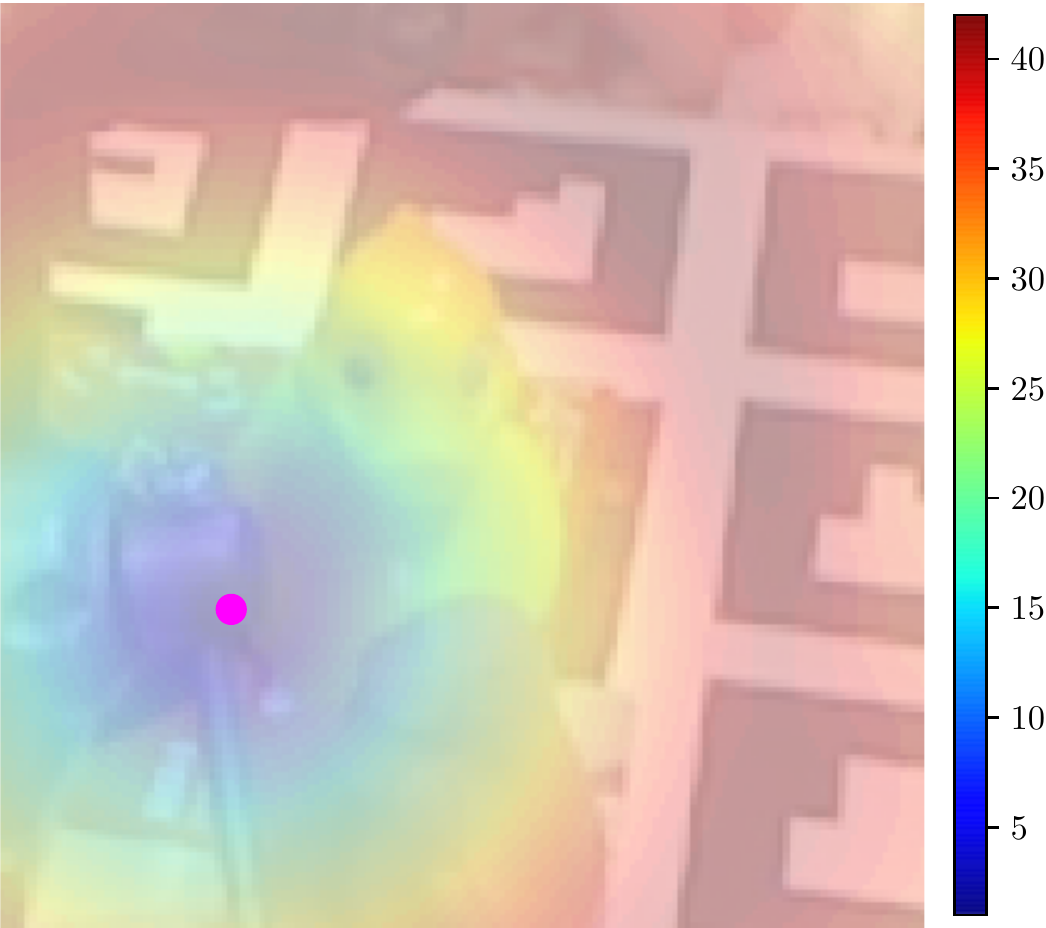}
}
\subfloat{%
    \includegraphics[height=\vizheight\linewidth, valign=t]{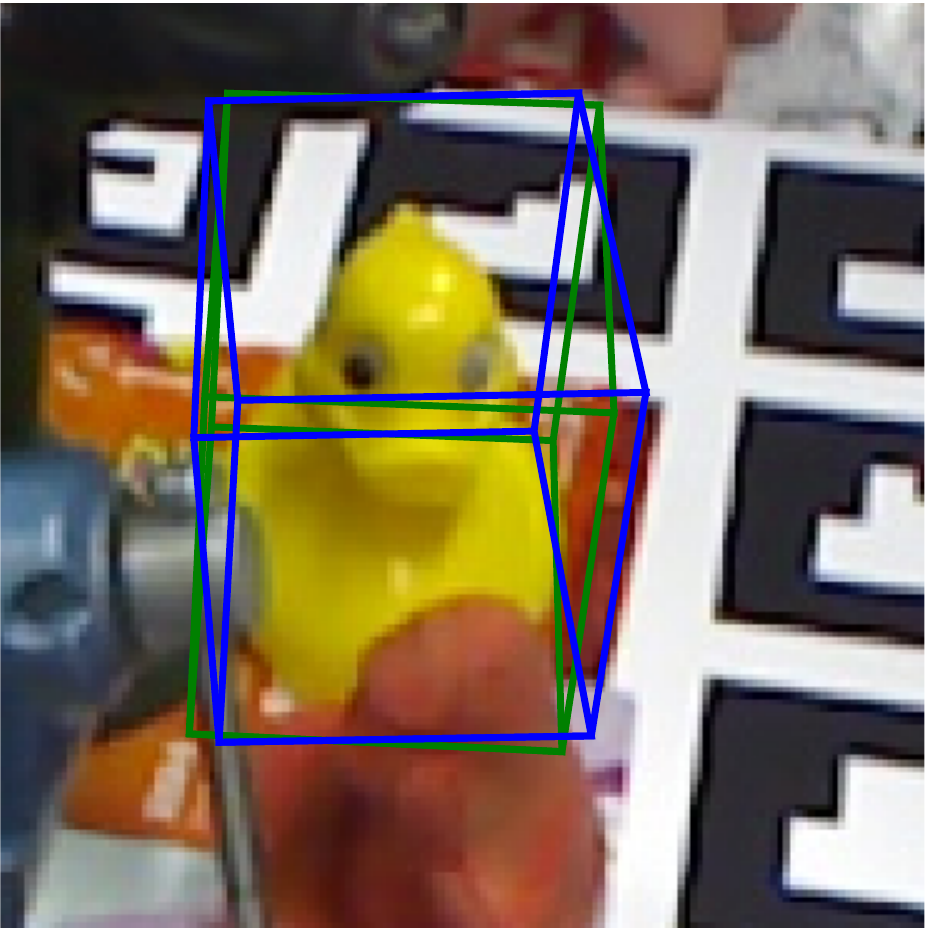}
}
\\
\subfloat{%
    \includegraphics[height=\vizheight\linewidth, valign=t]{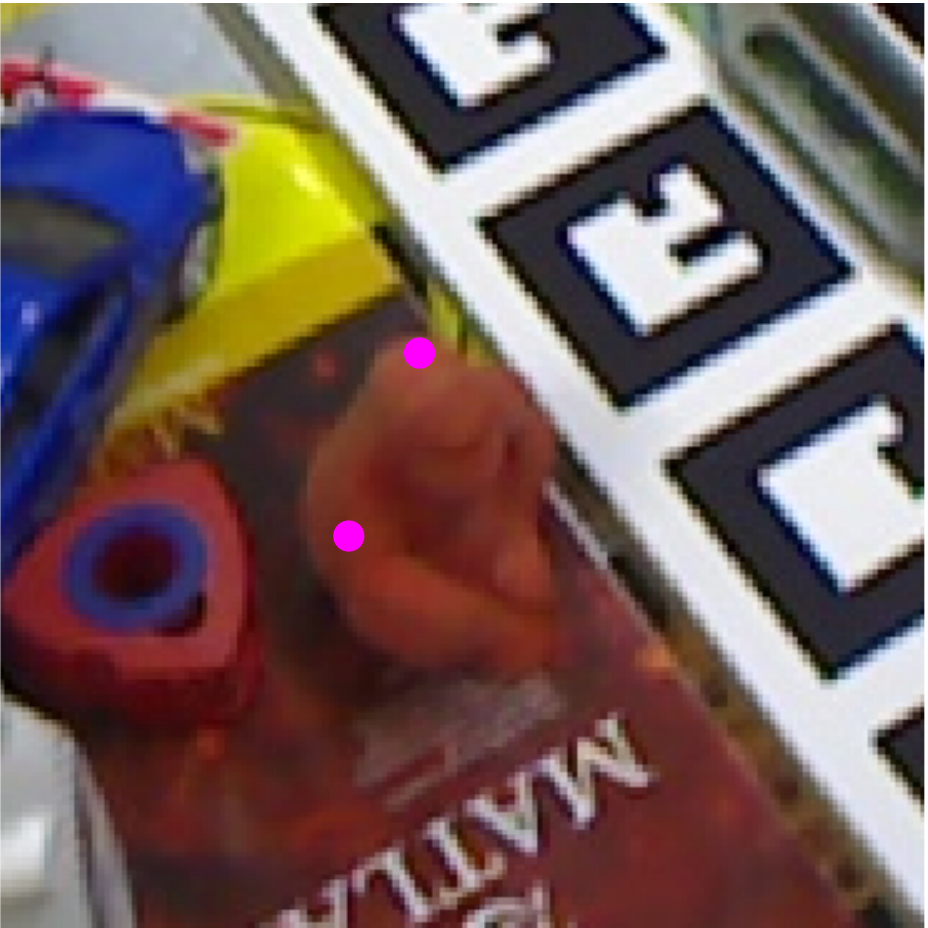}
}
\hspace{0.4ex}
\subfloat{%
    \includegraphics[height=\vizheight\linewidth, valign=t]{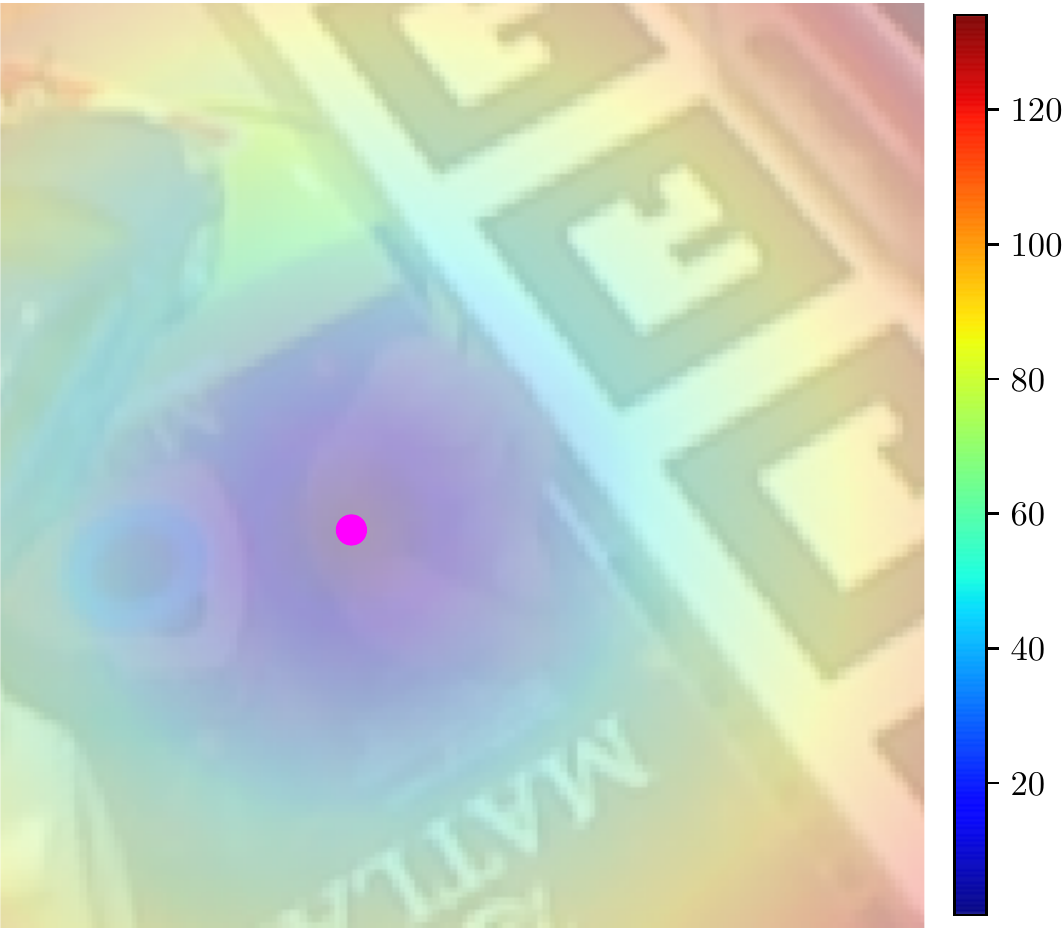}
}
\subfloat{%
    \includegraphics[height=\vizheight\linewidth, valign=t]{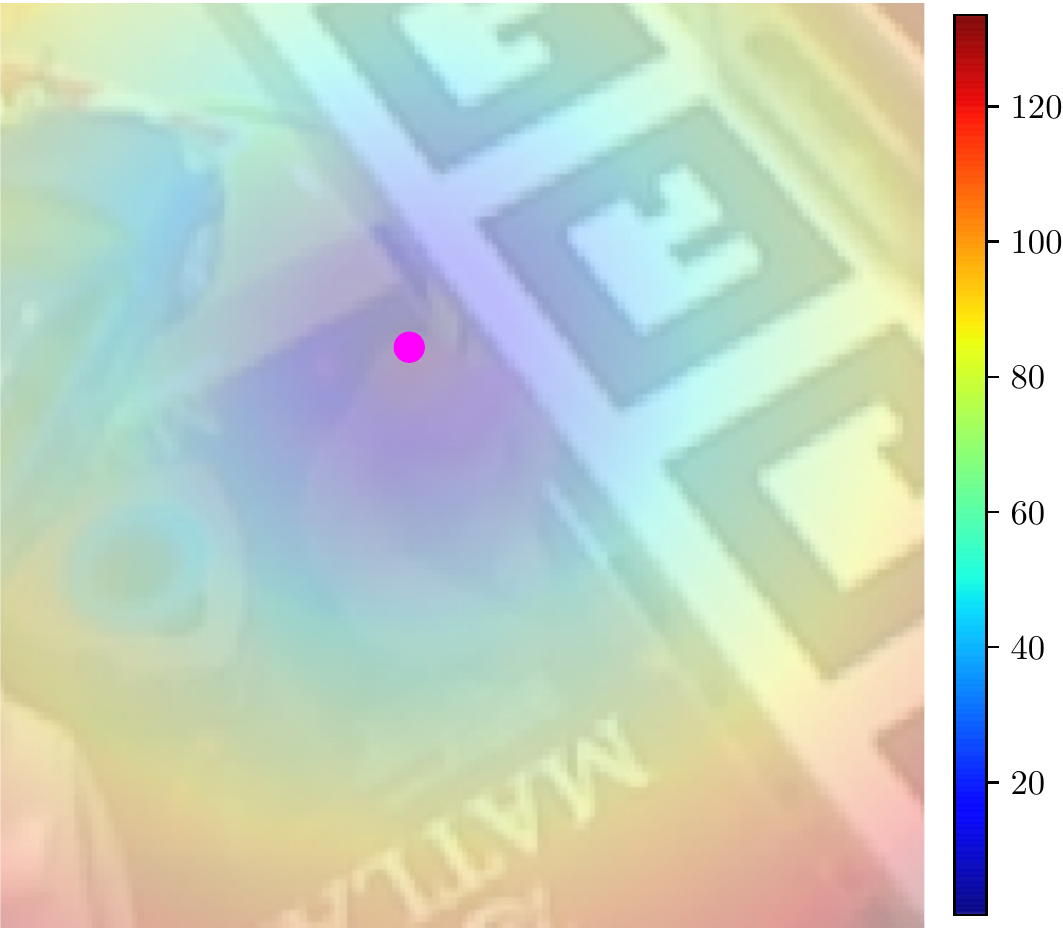}
}
\subfloat{%
    \includegraphics[height=\vizheight\linewidth, valign=t]{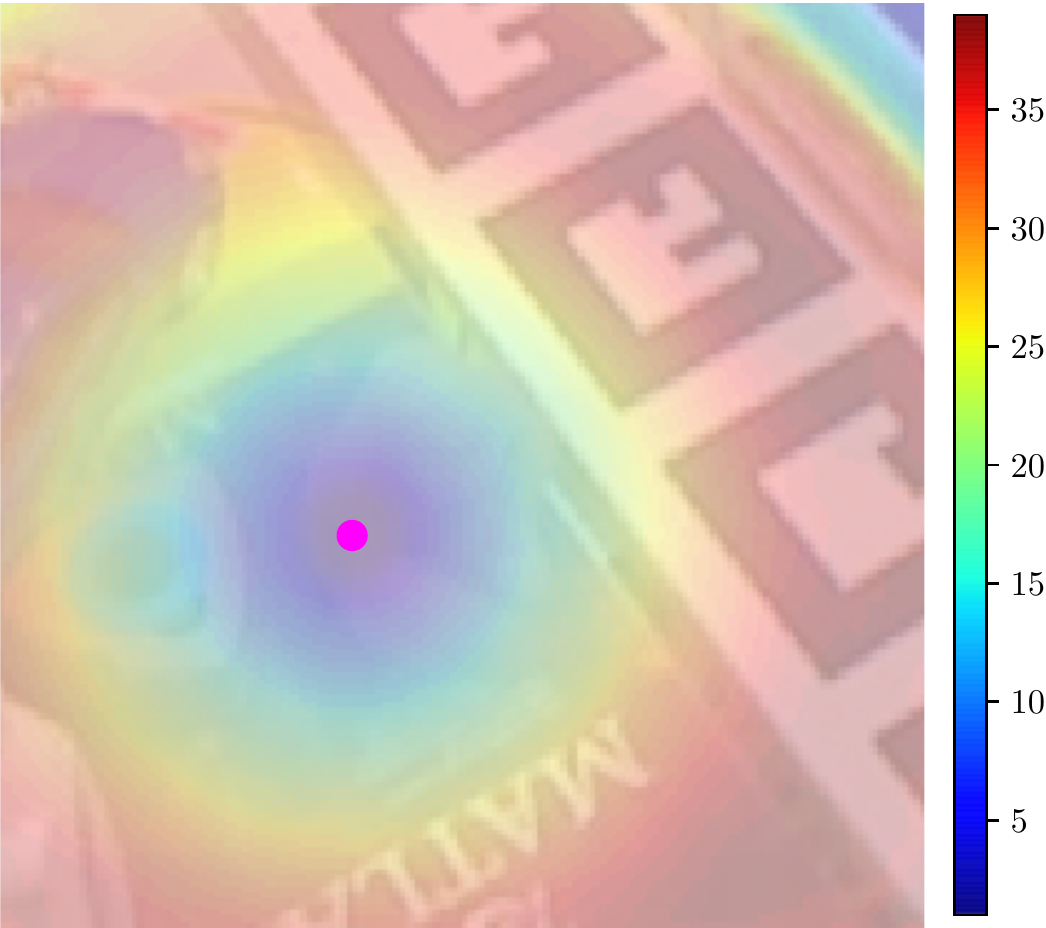}
}
\subfloat{%
    \includegraphics[height=\vizheight\linewidth, valign=t]{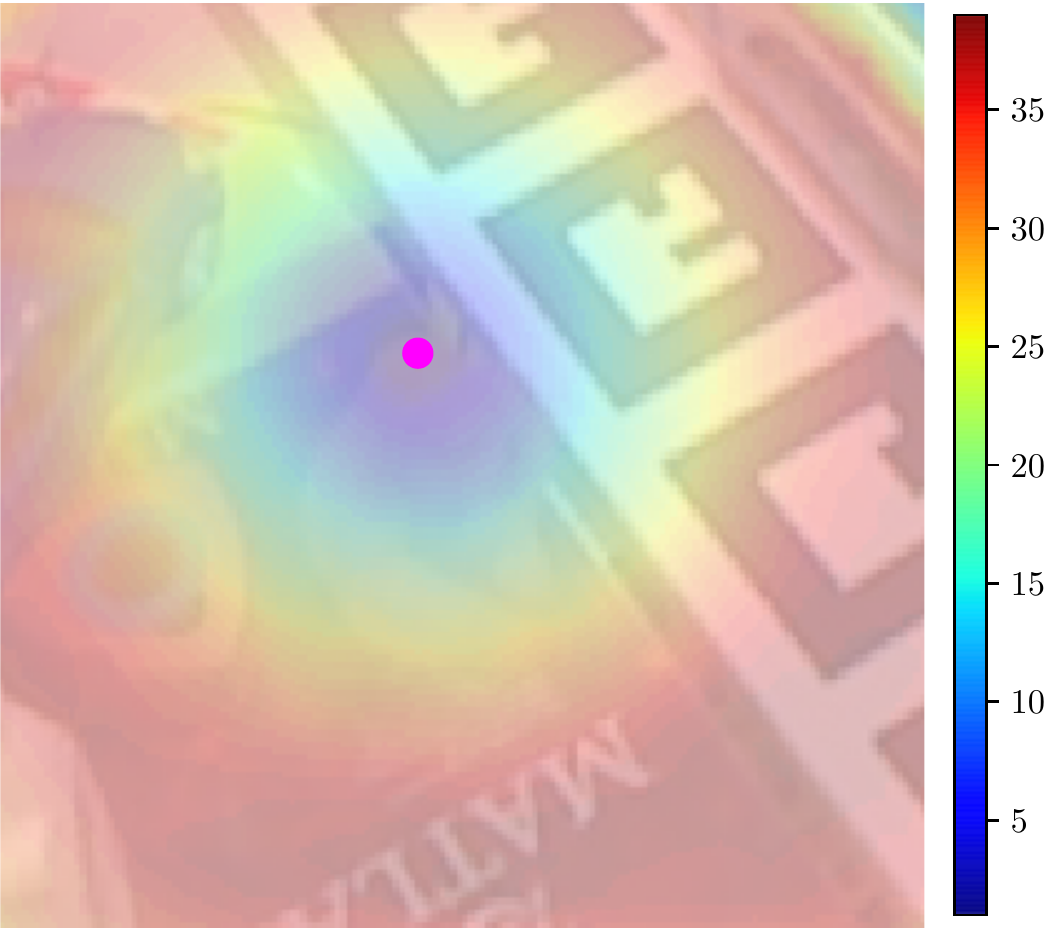}
}
\subfloat{%
    \includegraphics[height=\vizheight\linewidth, valign=t]{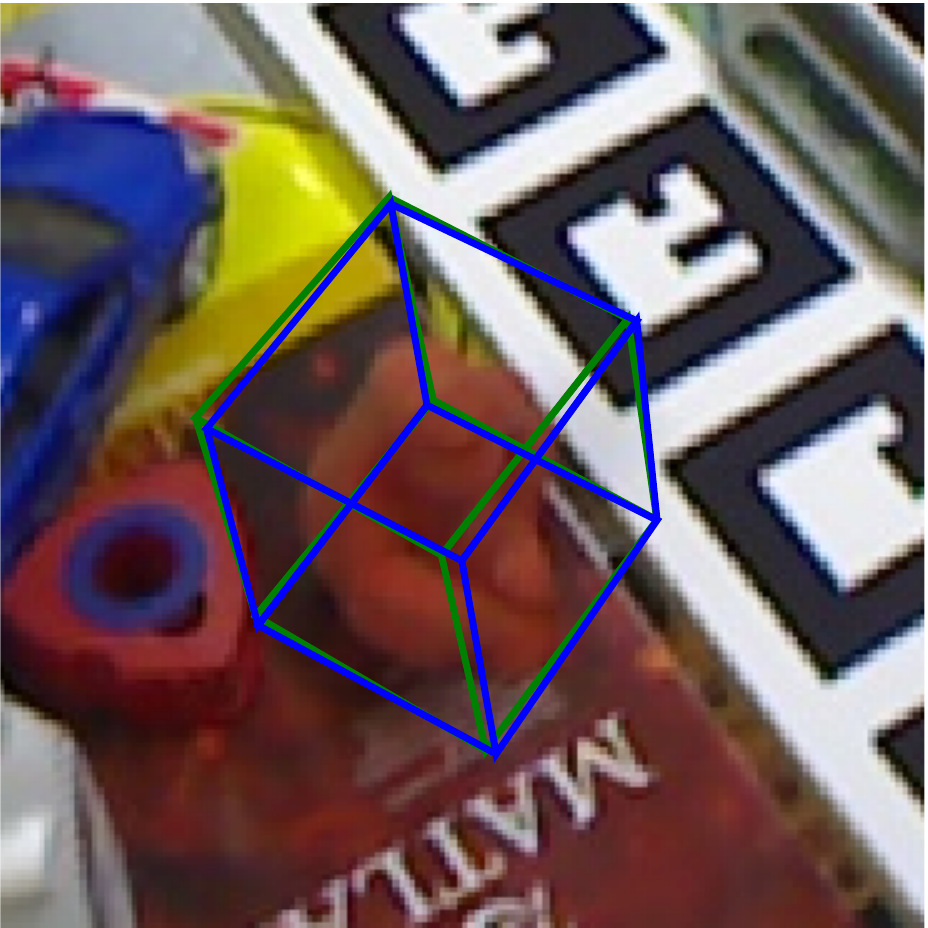}
}
\\
\subfloat{%
    \includegraphics[height=\vizheight\linewidth, valign=t]{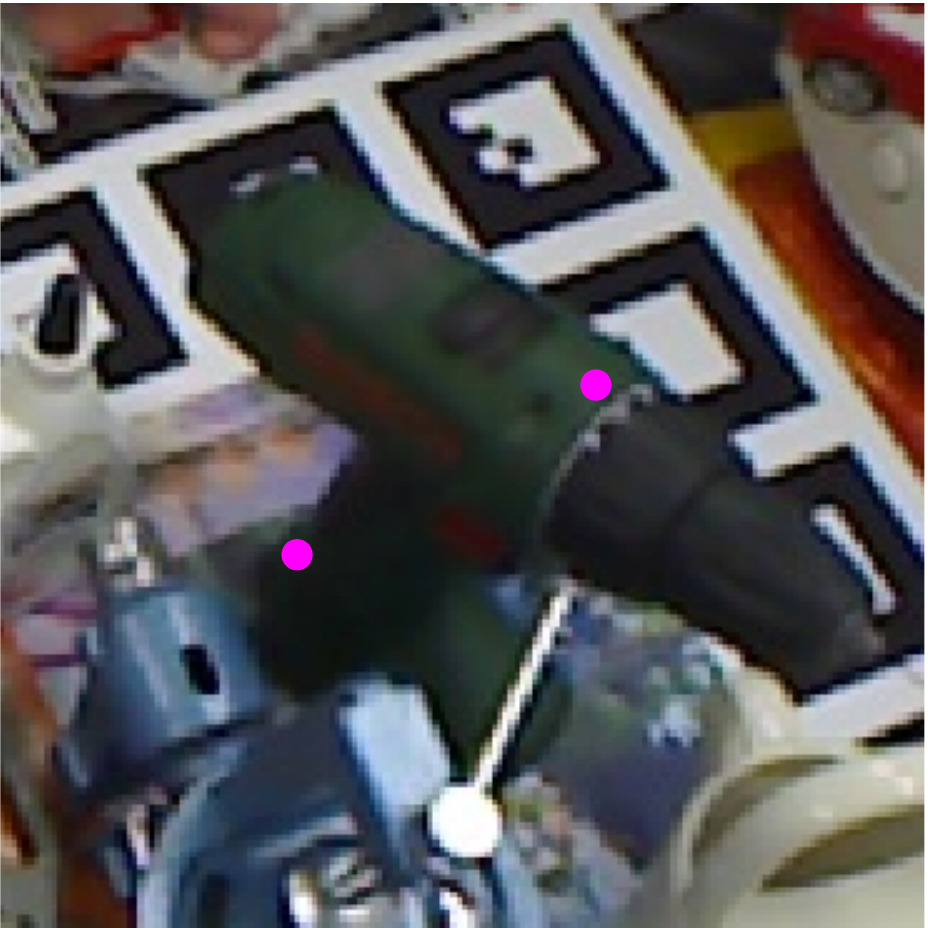}
}
\hspace{0.4ex}
\subfloat{%
    \includegraphics[height=\vizheight\linewidth, valign=t]{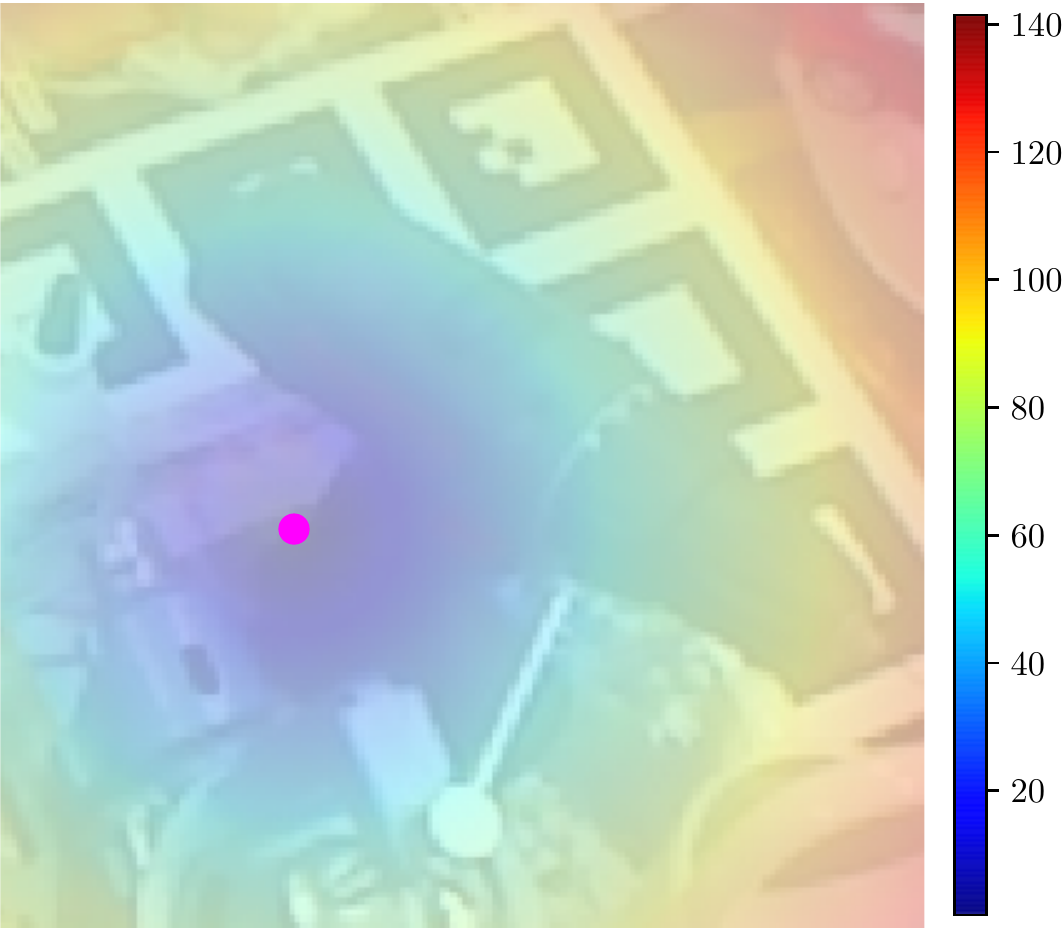}
}
\subfloat{%
    \includegraphics[height=\vizheight\linewidth, valign=t]{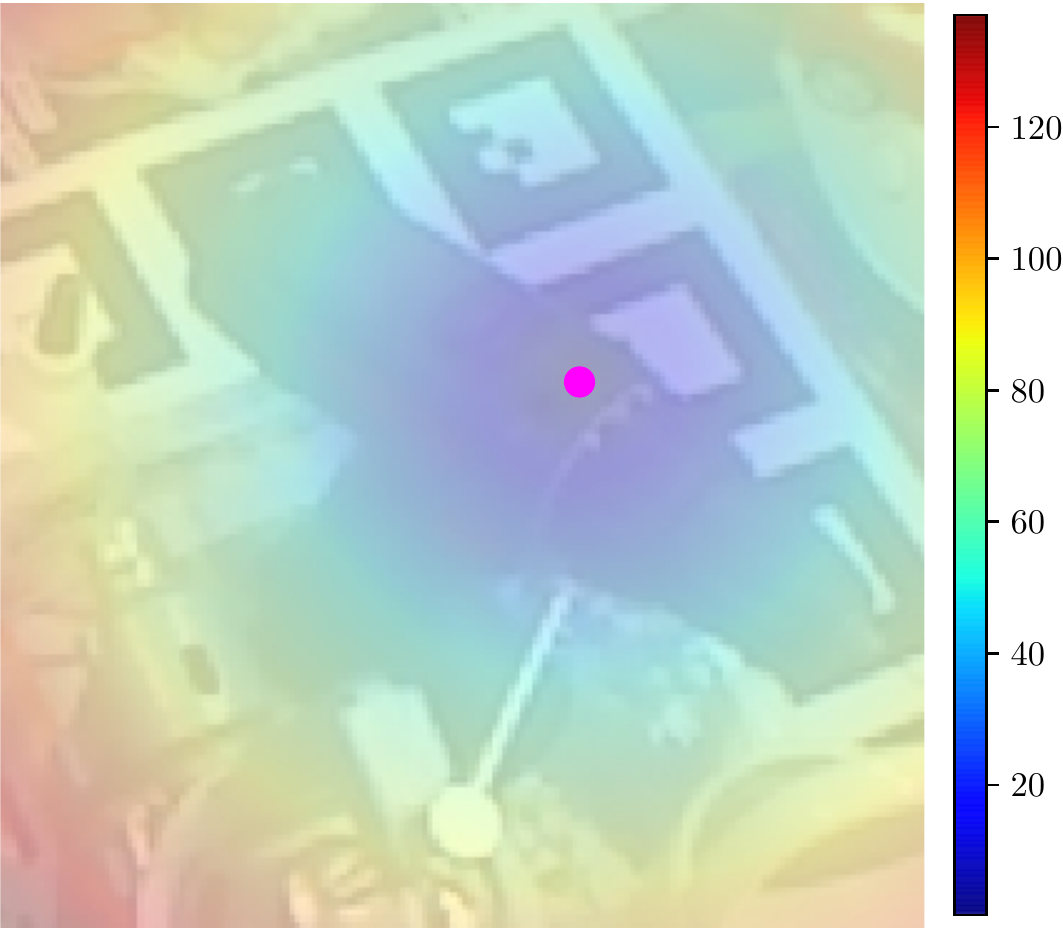}
}
\subfloat{%
    \includegraphics[height=\vizheight\linewidth, valign=t]{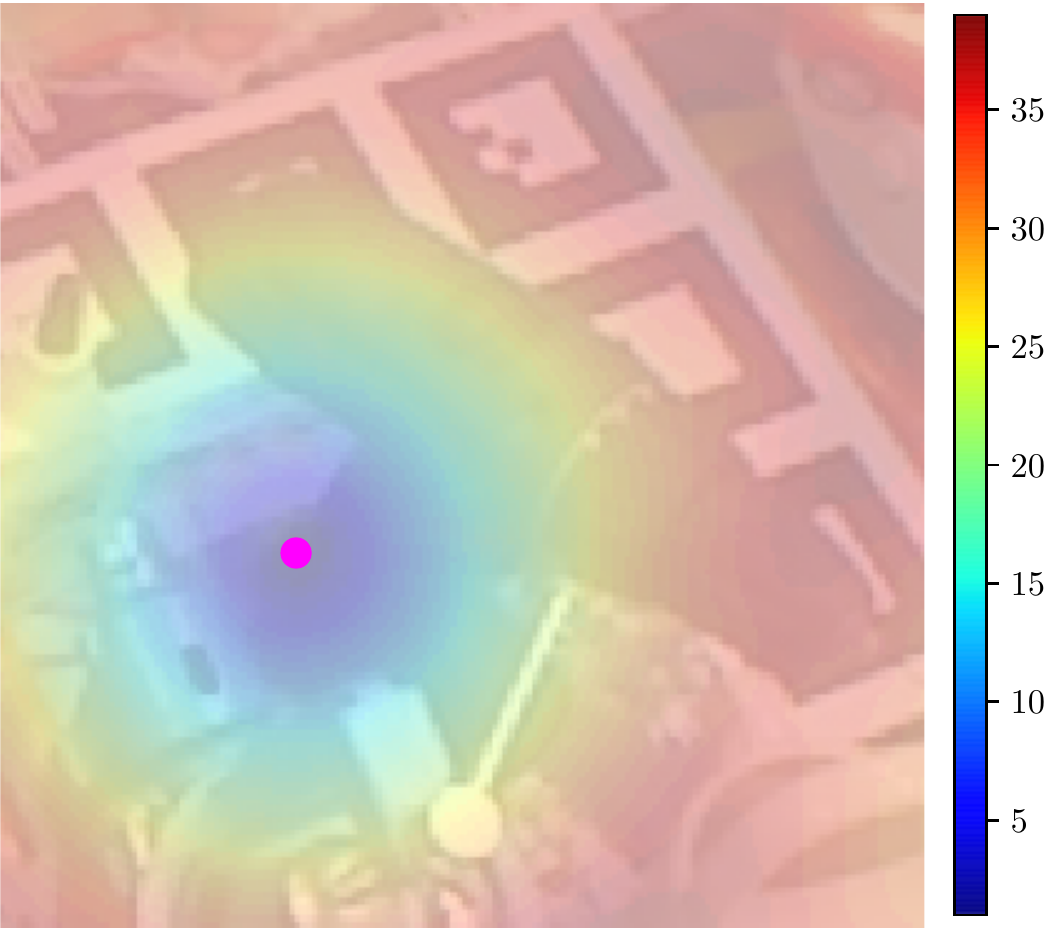}
}
\subfloat{%
    \includegraphics[height=\vizheight\linewidth, valign=t]{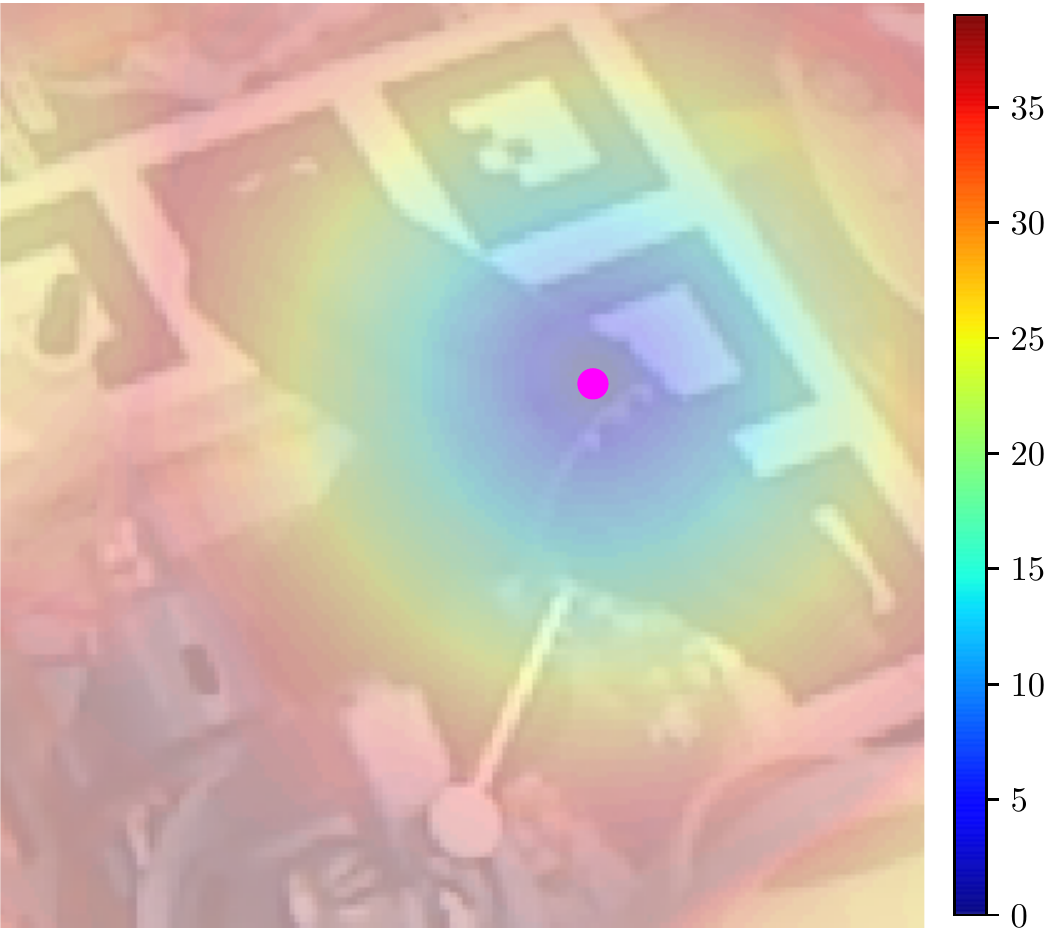}
}
\subfloat{%
    \includegraphics[height=\vizheight\linewidth, valign=t]{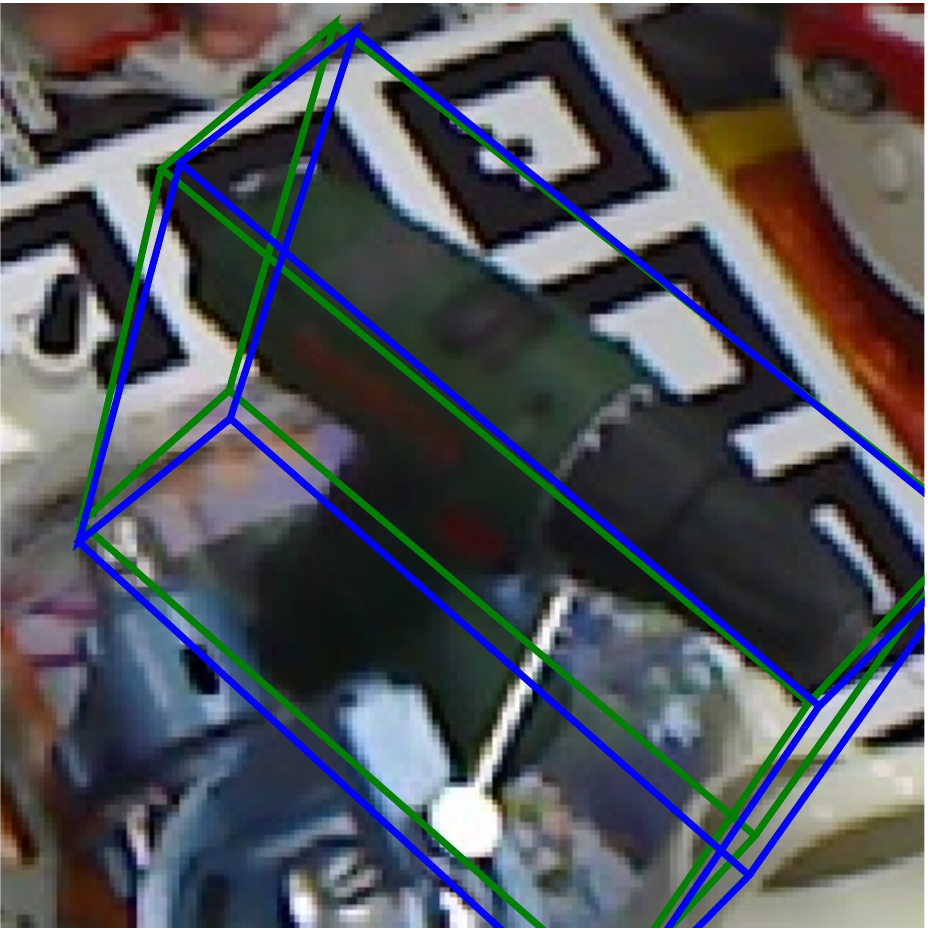}
}
\\
\subfloat{%
    \includegraphics[height=\vizheight\linewidth, valign=t]{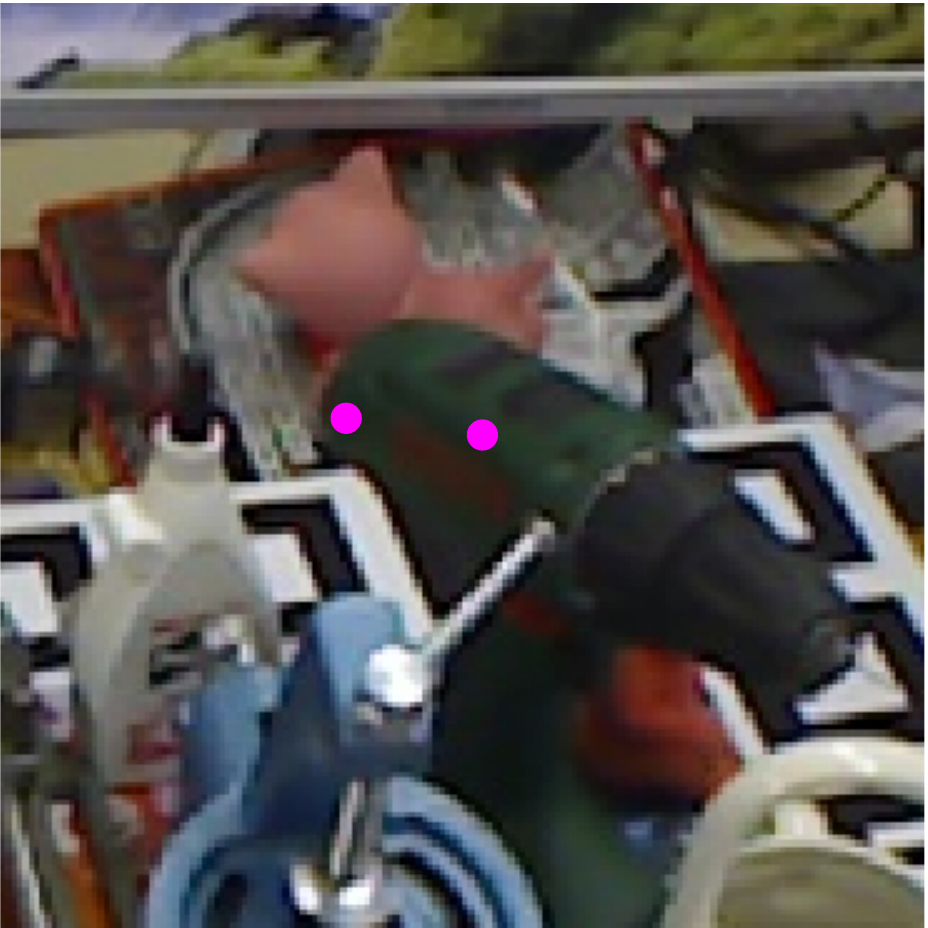}
}
\hspace{0.4ex}
\subfloat{%
    \includegraphics[height=\vizheight\linewidth, valign=t]{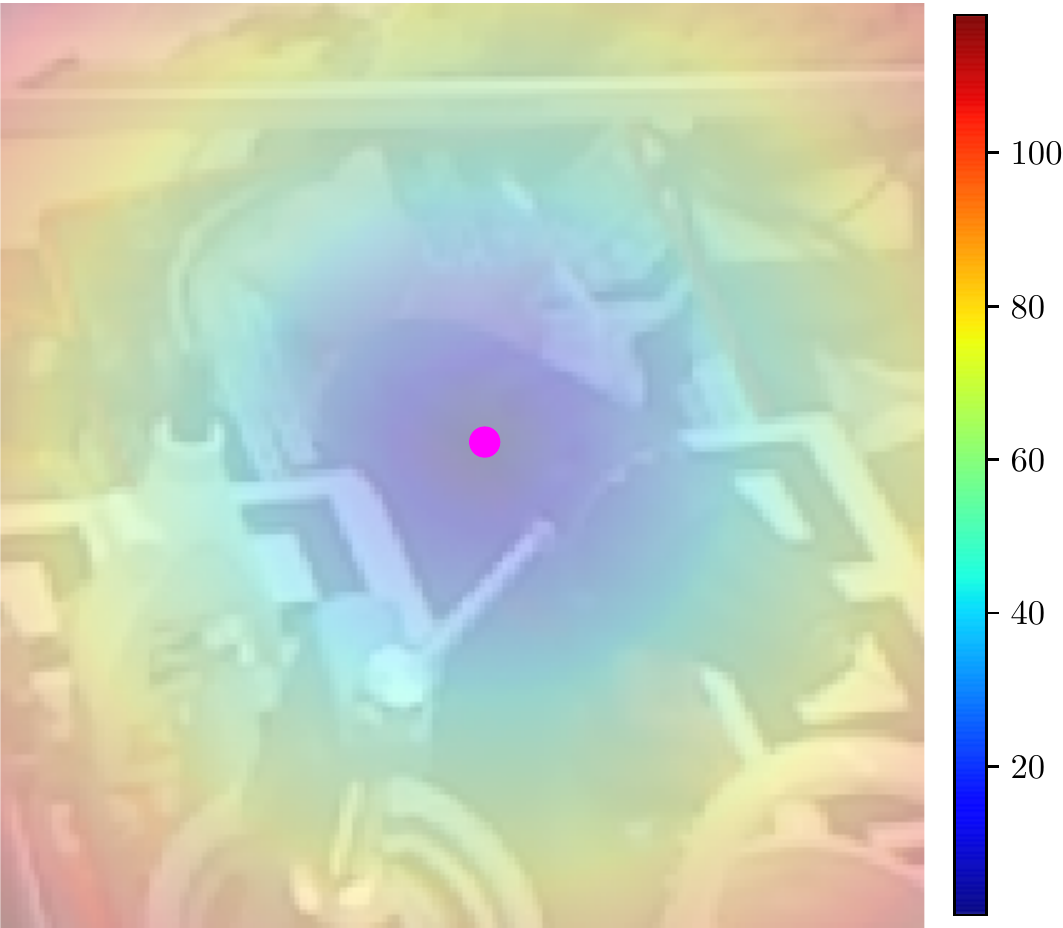}
}
\subfloat{%
    \includegraphics[height=\vizheight\linewidth, valign=t]{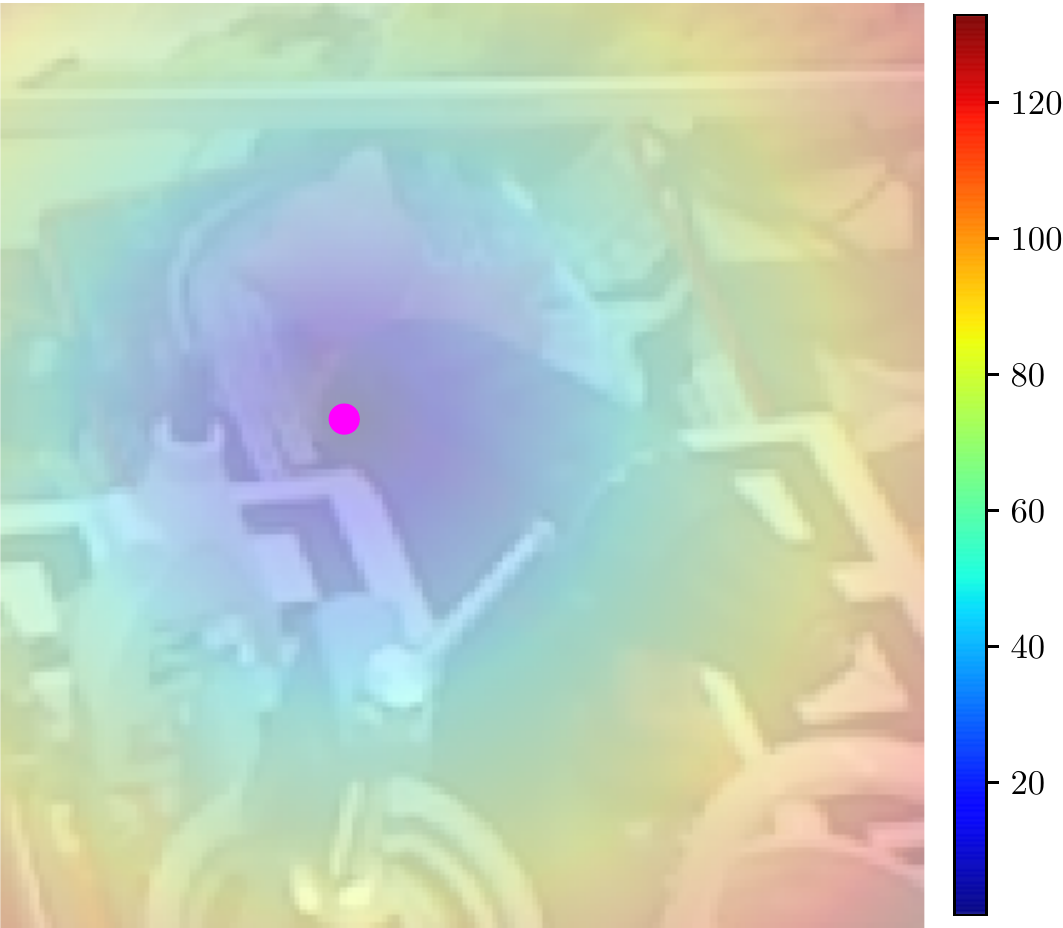}
}
\subfloat{%
    \includegraphics[height=\vizheight\linewidth, valign=t]{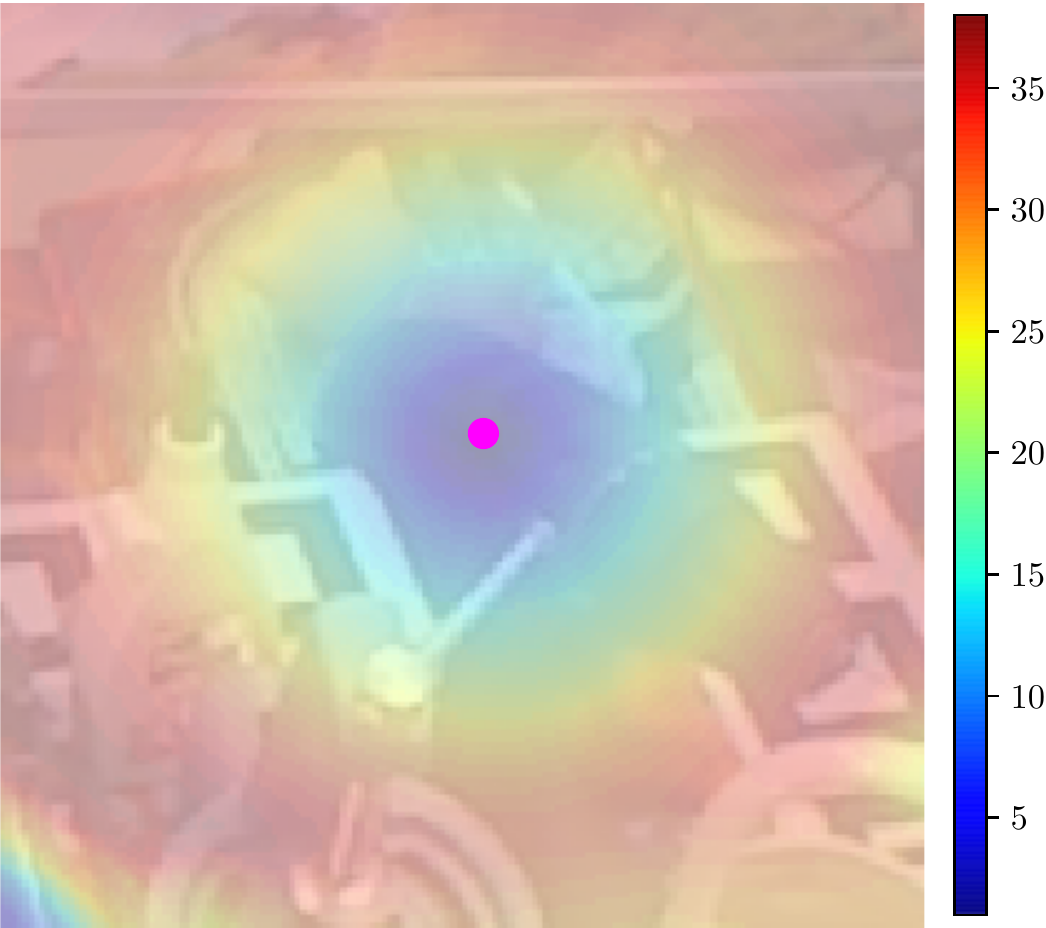}
}
\subfloat{%
    \includegraphics[height=\vizheight\linewidth, valign=t]{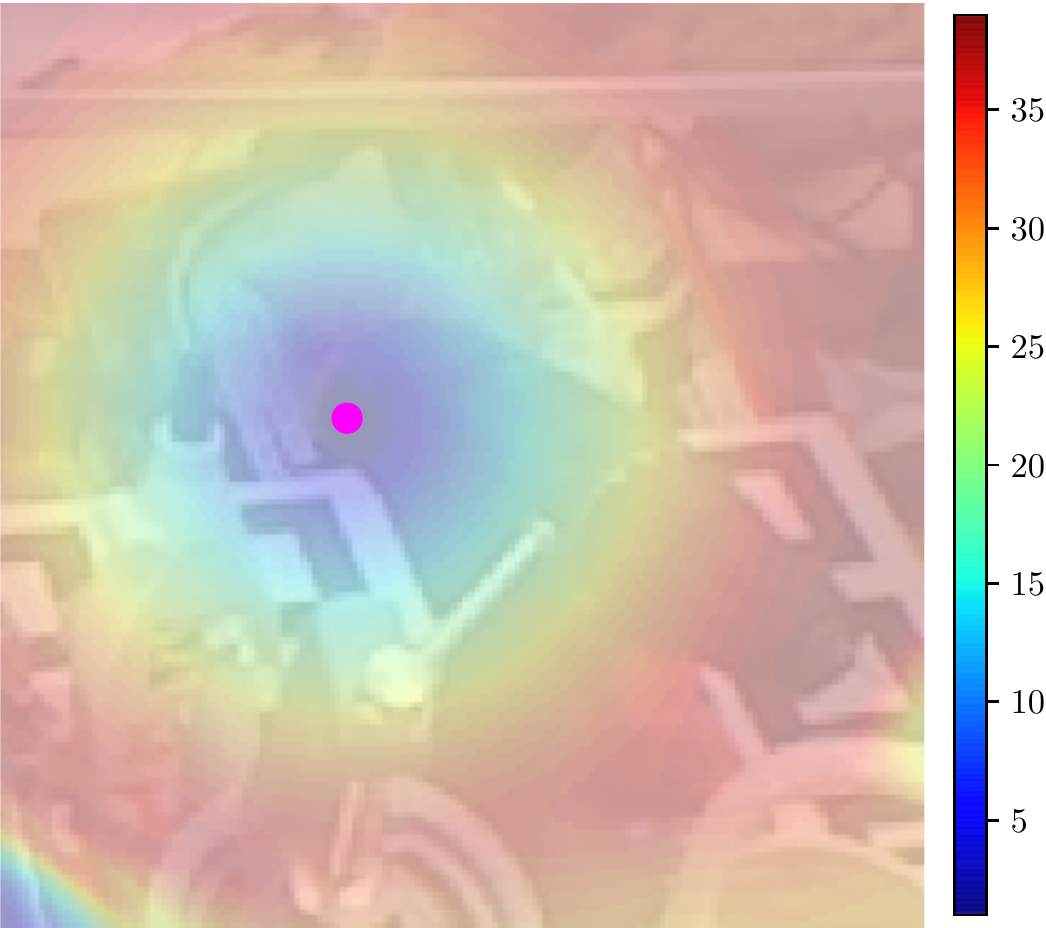}
}
\subfloat{%
    \includegraphics[height=\vizheight\linewidth, valign=t]{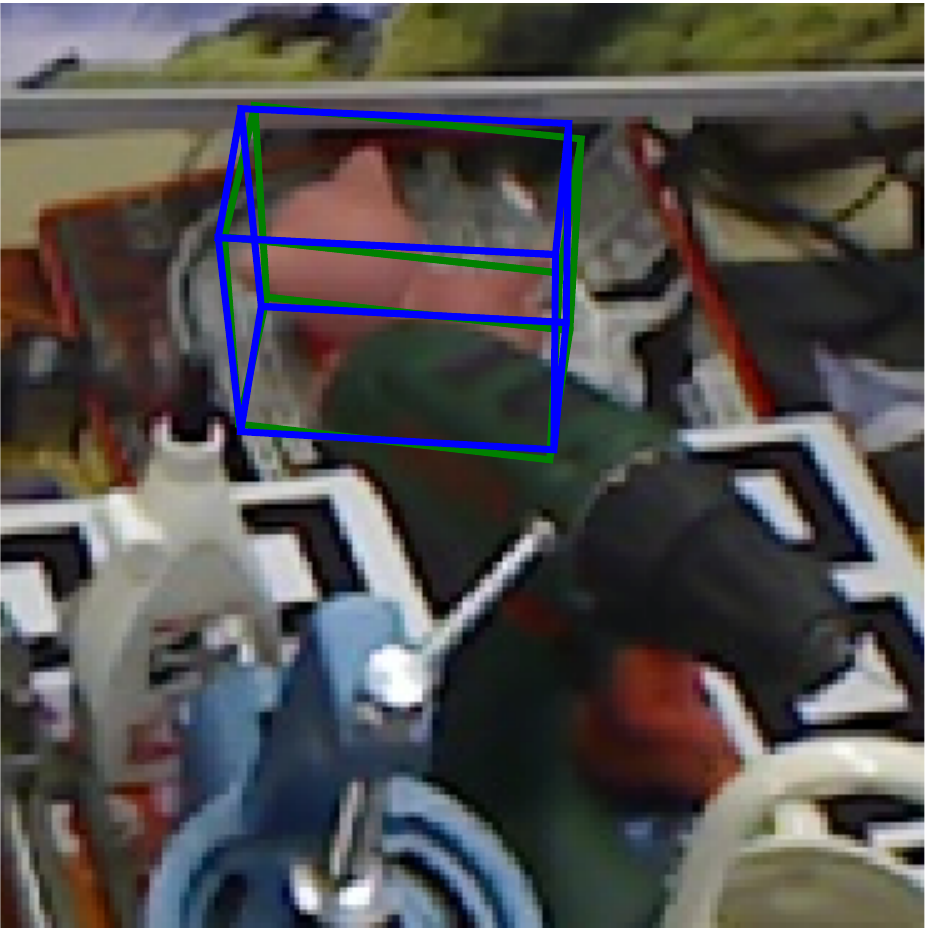}
}
\\
\subfloat{%
    \includegraphics[height=\vizheight\linewidth, valign=t]{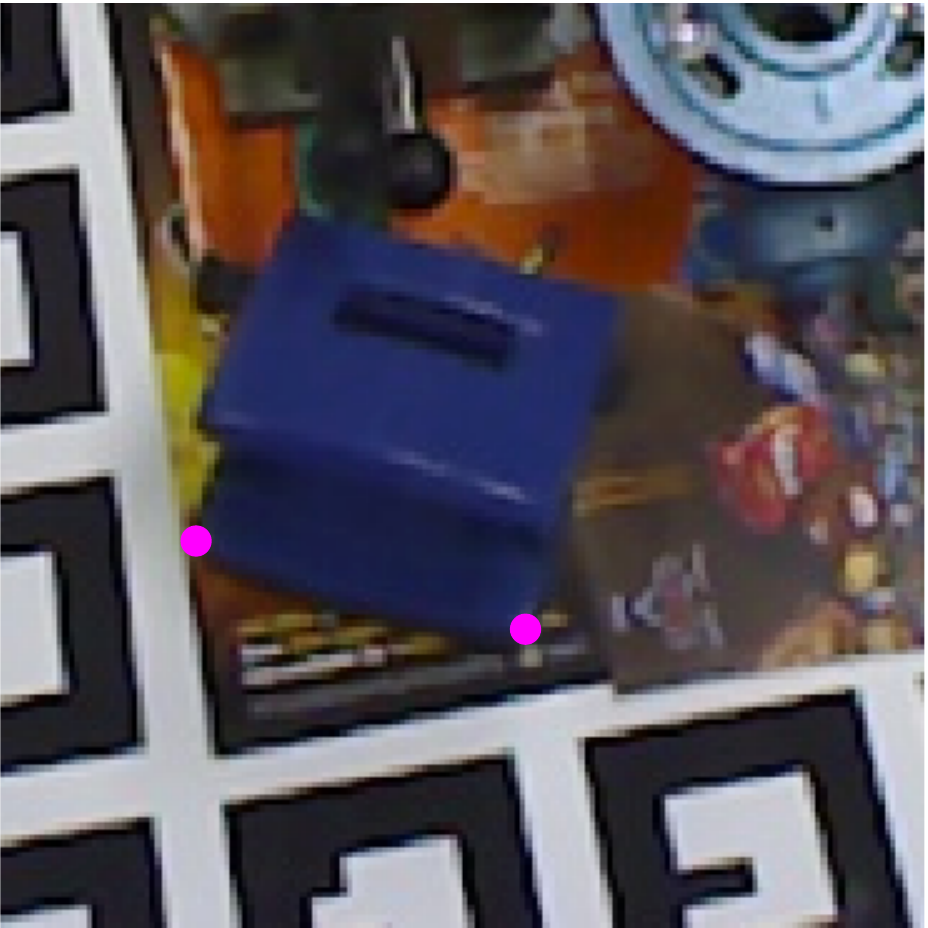}
}
\hspace{0.4ex}
\subfloat{%
    \includegraphics[height=\vizheight\linewidth, valign=t]{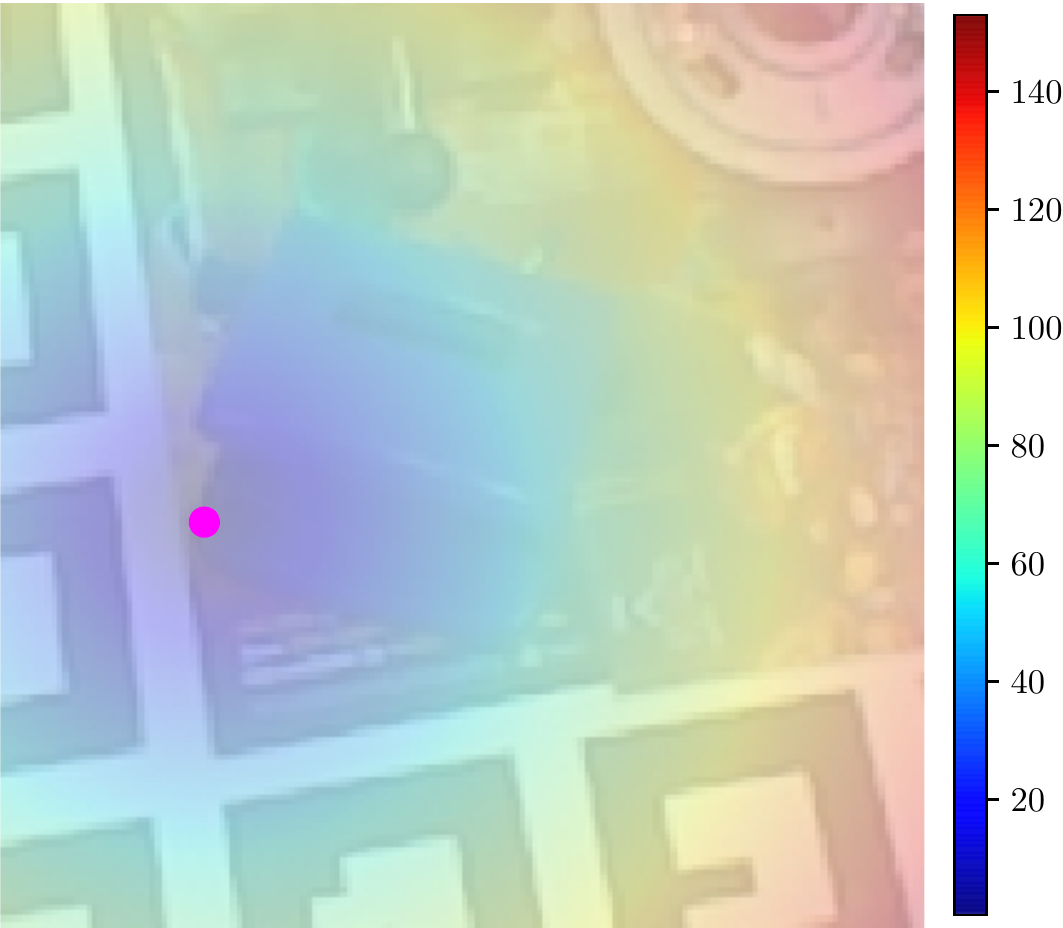}
}
\subfloat{%
    \includegraphics[height=\vizheight\linewidth, valign=t]{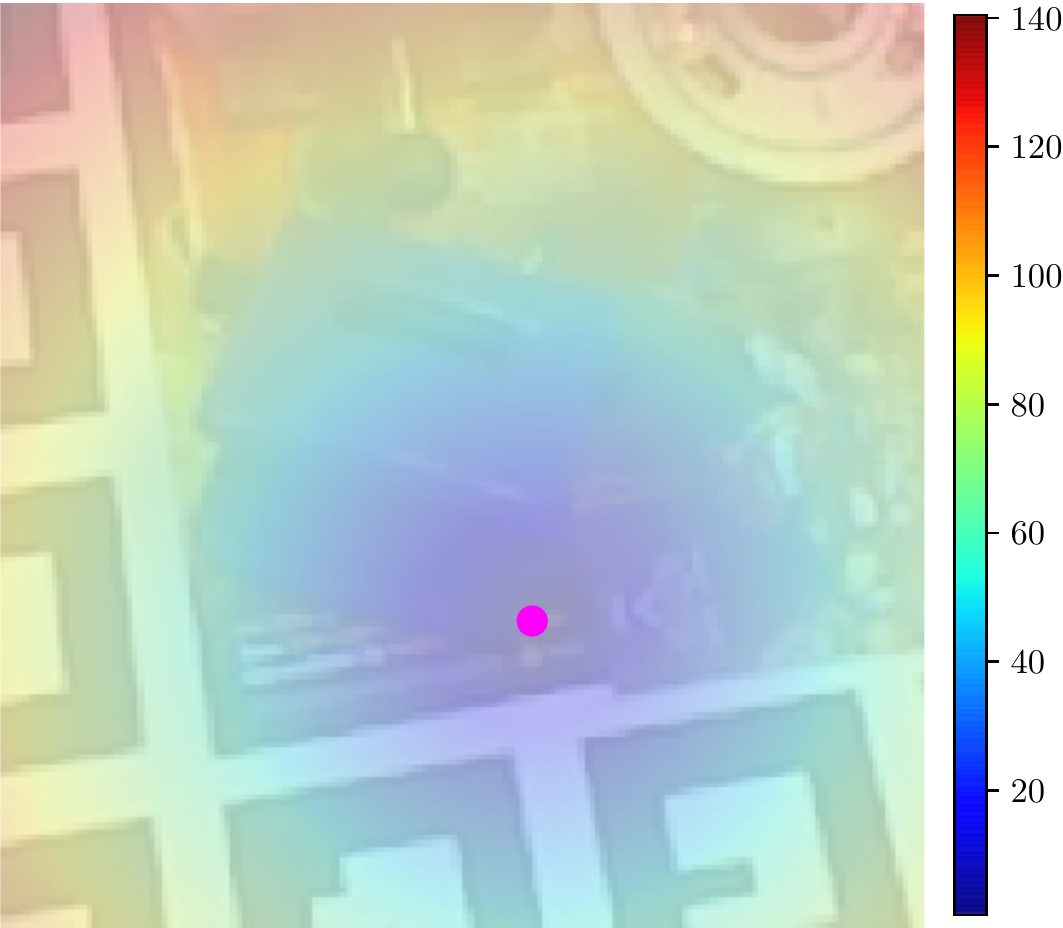}
}
\subfloat{%
    \includegraphics[height=\vizheight\linewidth, valign=t]{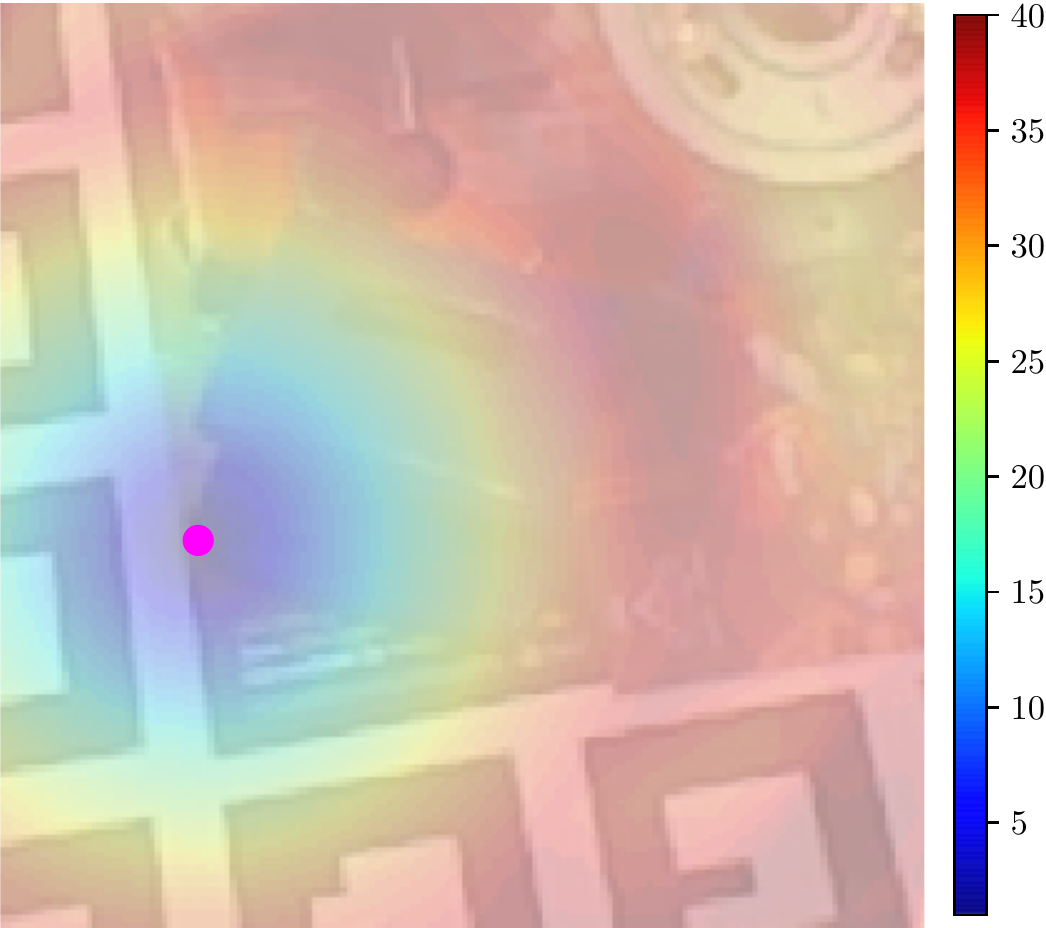}
}
\subfloat{%
    \includegraphics[height=\vizheight\linewidth, valign=t]{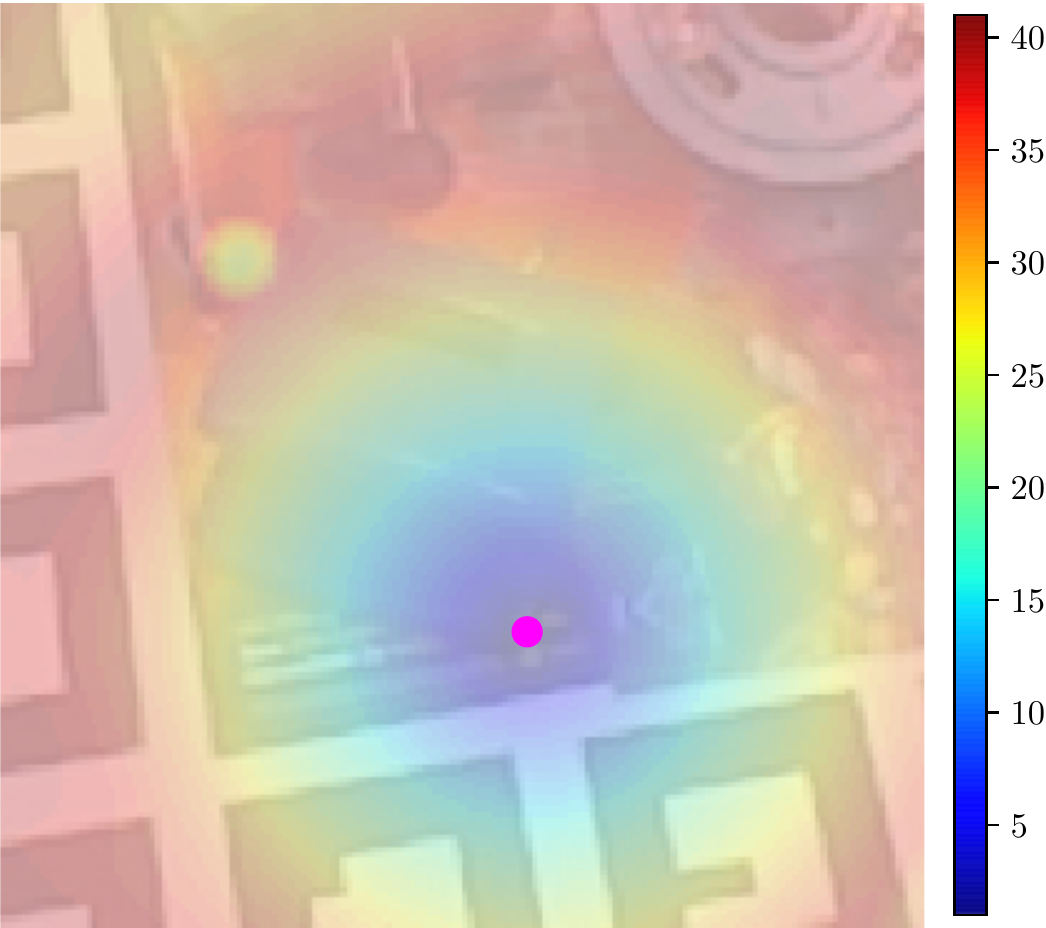}
}
\subfloat{%
    \includegraphics[height=\vizheight\linewidth, valign=t]{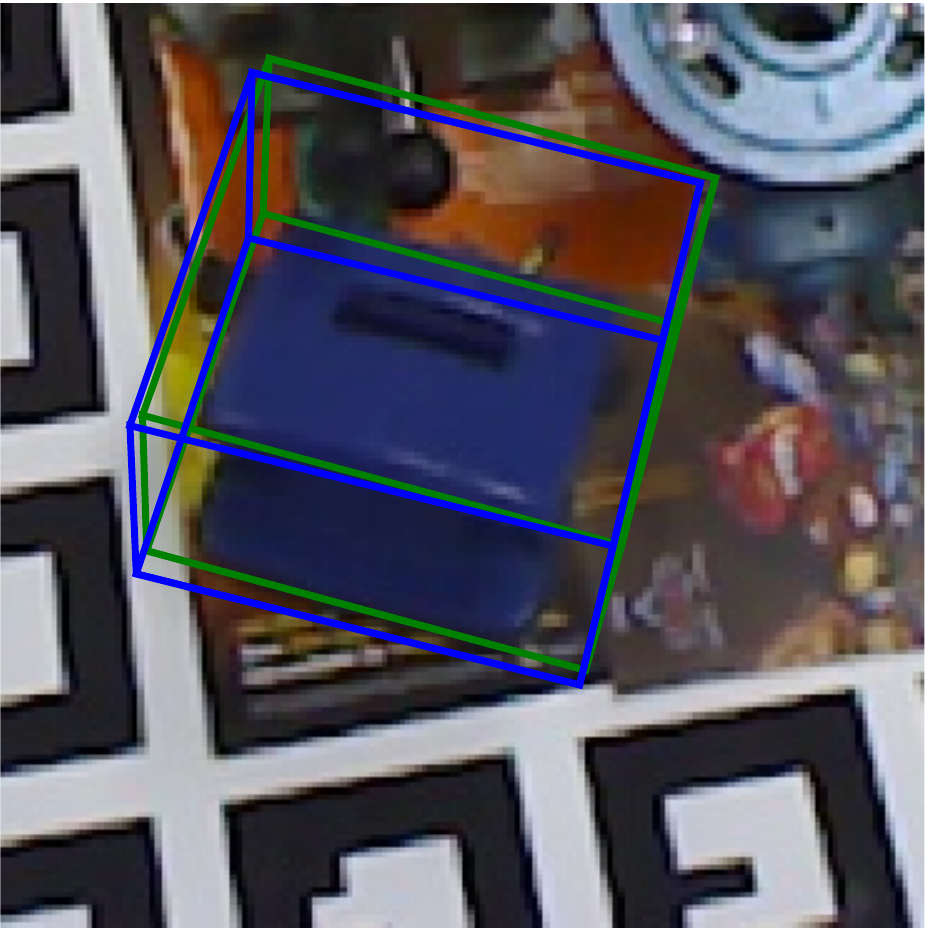}
}
\caption{\textbf{Visualization of predicted keypoints, distance fields and 6D poses} on Occlusion LINEMOD dataset.
From left to right in each row: (a) input image with two of the ground truth keypoints; (b)(c) two ground truth distance fields and locations of the keypoints; (d)(e) two predicted distance fields and voted location of the keypoints; (f) ground truth 6D pose (green 3D bounding box) and predicted 6D pose (blue 3D bounding box).
}
\label{fig:viz:map}
\vspace{-1ex}
\end{figure*}

{\small
\bibliographystyle{ieeetr}
\bibliography{bib}
}

\end{document}